\documentclass[conference]{IEEEtran}
\IEEEoverridecommandlockouts
\usepackage{cite}
\usepackage{amsmath,amssymb,amsfonts, mathtools}
\usepackage{algorithm}
\usepackage{algpseudocode}
\usepackage{graphicx}
\usepackage{textcomp}
\usepackage{xcolor}
\usepackage{booktabs}
\usepackage{subcaption}
\usepackage{multirow}
\def\BibTeX{{\rm B\kern-.05em{\sc i\kern-.025em b}\kern-.08em
    T\kern-.1667em\lower.7ex\hbox{E}\kern-.125emX}}

\author{Ayse Gul Yildiz$^{1}$, Rekha Raja$^{*1}$
\thanks{$^{1}$School of Artificial Intelligence, Donders Centre for Cognition, Department of Human-Centred Intelligent Systems, Radboud University, Nijmegen, The Netherlands.}
\thanks{*Email: rekha.raja@donders.ru.nl}}

\begin{document}

\title{Receding-Horizon Next-Best-View Planning for Highly Occluded Target Perception }

\maketitle

\begin{abstract}
Robotic systems that must acquire information about a specific target often face severe occlusion from surrounding structures or from the target itself, such that a single camera viewpoint provides only partial information. Active perception must therefore select viewpoints that progressively reveal unobserved regions. However, conventional local next-best-view (NBV) methods typically select viewpoints according to their immediate utility, which can overlook views that provide limited information initially but enable more informative observations from subsequent viewpoints. We formulated local target perception as a short-horizon sequential view-planning problem and proposed a Receding-Horizon Next-Best-View (RH-NBV) planner that evaluates candidate sequences of future viewpoints rather than isolated views. 
RH-NBV is evaluated on a UR5e manipulator with an eye-in-hand RGB-D camera in simulation as well as on physical robot against a gradient-based NBV planner, and random search, using a coffee mug and Stanford bunny under controlled occlusion and a tomato plant with natural self-occlusion. Under heavy occlusion, RH-NBV achieves substantially higher target information recovery, reaching 63.8\% ROI coverage versus 34.2\% and 85.5\% occluded-surface recall versus 7.4\% for the single-step NBV baselines on the most occluded mug configuration. A controlled horizon ablation shows that this improvement arises from multi-step reasoning rather than additional computation. RH-NBV provides the greatest benefit for strongly occluded targets, while myopic NBV remains competitive and substantially cheaper when visibility is less constrained.

\end{abstract}


\section{Introduction}
\label{sec:intro}

Active perception enables a robot to acquire information that cannot be obtained from a fixed sensor configuration \cite{kootstra_wang_blok_hemming_henten_2021}. By moving its sensor to different viewpoints, a robot can improve the visibility of task-relevant regions and acquire complementary observations of the environment \cite{bajcsy_1988, aloimonos_weiss_bandyopadhyay_1988}. This capability is particularly important when a target is partially occluded by surrounding structures or by itself, as in object inspection, manipulation, harvesting, and active 3D perception. 
Next-best-view (NBV) planning provides a framework for actively selecting such viewpoints according to their expected utility. While many NBV approaches address global scene exploration or reconstruction, local target-centric perception requires sensing to be focused on a specific region of interest (ROI). 
Under occlusion, however, viewpoint utility can be sequential: a view with limited immediate information may reveal regions that enable more informative subsequent observations. Conventional local NBV methods typically select the viewpoint with the highest immediate utility and therefore do not explicitly account for these downstream effects.

We formulated local target perception under occlusion as a short-horizon sequential view-planning problem and proposed a Receding-Horizon Next-Best-View (RH-NBV) planner. RH-NBV samples candidate sequences of future viewpoints and evaluates each sequence using a semantic, occlusion-aware utility while propagating the belief state through the sequence. The planner executes only the first viewpoint of the selected sequence, incorporates the acquired observation into the belief, and recomputes the planning horizon. This formulation enables the planner to account for the future visibility that a viewpoint can create while retaining the feedback and adaptability of conventional single-step planning. We evaluate RH-NBV on a UR5e manipulator with an eye-in-hand RGB-D camera against a gradient-based NBV planner, and random search. Experiments use a coffee mug and the Stanford bunny under a controlled six-stage occlusion ladder, as well as a tomato plant exhibiting natural self-occlusion. 
The results show that RH-NBV provides its greatest advantage under strong occlusion, while myopic NBV remains competitive and substantially cheaper when the target is more readily visible.

The paper makes several contributions: (i) a formulation of local target perception under occlusion as a short-horizon sequential NBV problem, where a viewpoint depends on the future visibility it enables; (ii) RH-NBV combines semantic target information, occlusion-aware utility, and in-horizon belief-state propagation; (iii) an evaluation across controlled and natural occlusion conditions showing that the benefit of non-myopic planning increases as target visibility becomes more constrained, (iv) implemented the method on a physical systems and showed the validation and effectiveness of the proposed method.

\section{Related Work}
\label{sec:related}

The next-best-view (NBV) problem was introduced for automated 3D reconstruction by Connolly \cite{connolly_1985}, who iteratively selected viewpoints expected to expose the largest amount of previously unmapped volume. Subsequent approaches have extended NBV using information gain, visibility, uncertainty, and geometric reconstruction objectives. A common characteristic is that candidate viewpoints are evaluated with respect to the current reconstruction state and the highest-utility viewpoint is selected for execution. Such formulations are therefore \emph{myopic}: they optimize immediate observations without considering how a viewpoint affects future information, which is particularly limiting under occlusion.

Receding-horizon approaches address this limitation by evaluating multiple future viewpoints before executing the next action. Bircher \emph{et al.} \cite{bircher_kamel_alexis_oleynikova_siegwart_2016} introduced a receding-horizon NBV strategy for autonomous exploration of unknown 3D environments, constructing a tree of collision-free future configurations and selecting branches according to expected unmapped volume and path cost. Zhang and Zhang \cite{zhang_zhang_2022} similarly applied receding-horizon planning to active 3D reconstruction with a manipulator, evaluating candidate sequences using occupancy entropy accumulated along their views. These methods demonstrate the value of non-myopic view planning, but primarily address global exploration or reconstruction rather than information acquisition focused on a semantically defined target whose visibility evolves throughout the sequence.

Occlusion-aware and target-specific NBV provide complementary capabilities. Delmerico \emph{et al.} \cite{delmerico2018} incorporated visibility into volumetric information gain by weighting voxel uncertainty according to ray reachability. 
For target-centric reconstruction, restricting the objective to a task-relevant ROI can prevent sensing effort from being spent on irrelevant scene geometry \cite{zaenker_claus_mccool_bennewitz_2020}. 
GradientNBV \cite{burusa_henten_gert_kootstra_2024} is particularly closely related to this work. It combines semantic target information with an occlusion-aware information objective for local perception of plant nodes, representing occupancy and semantic confidence in a voxel map and using gradient ascent to efficiently identify informative camera viewpoints. Its differentiable formulation achieved reconstruction performance comparable to a sampling-based semantic planner while substantially reducing ray-tracing evaluations. However, GradientNBV remains single-step: it optimizes the utility of the immediate viewpoint and does not explicitly account for the future information that a viewpoint may enable. This distinction is important under severe occlusion, where successive viewpoint utilities depend on information acquired earlier.

Together, prior work establishes the key components of informative active perception.
 
The remaining gap is their combination for local perception of a semantically defined target under severe occlusion, where viewpoint value depends on preceding observations. RH-NBV addresses this gap by evaluating short viewpoint sequences with semantic and occlusion-aware utility and predicted belief-state propagation through the horizon.

\section{Method}
\label{sec:method}

\subsection{Problem Formulation}
\label{sec:problem}
We consider a 6-DoF manipulator equipped with an eye-in-hand RGB-D camera operating in a bounded workspace $\mathcal{V}_T\subset\mathbb{R}^3$ containing a target object and potentially occluding structures. The objective is to acquire information about a target within a predefined region of interest (ROI), $\mathcal{V}_{{\mathrm{ROI}}^T}\subset\mathcal{V}_T$, while avoiding unnecessary sensing of the surrounding scene. The ROI location is assumed known, whereas the target geometry and occupancy are initially unknown. Semantic information from an instance-segmentation module identifies the target and distinguishes it from surrounding structures. Let $\mathcal{V}_C\subset\mathbb{R}^3$ denote the admissible camera-position space, with $\mathcal{V}_C\cap\mathcal{V}_T=\emptyset$. A viewpoint is parameterized by its camera position $\mathbf{p}\in\mathcal{V}_C$. Its orientation is constrained to point toward a fixed target point $\mathbf{t}\in\mathcal{V}_{{\mathrm{ROI}}^T}$ via $\mathbf{q}=\operatorname{LookAt}(\mathbf{p},\mathbf{t})$
so that viewpoint planning is reduced to the three-dimensional position $\mathbf{p}$. Camera roll is not explicitly optimized. At planning iteration $n$, let $\mathcal{S}_n$ denote the current belief state obtained from all observations collected so far. RH-NBV considers a sequence of $H$ future viewpoints, $\boldsymbol{\Xi}_n=[\mathbf{p}_{n,1},\ldots,\mathbf{p}_{n,H}]$, $\mathbf{p}_{n,h}\in\mathcal{V}_C$.

Starting from $\mathcal{S}_{n,0}=\mathcal{S}_n$, each hypothetical observation induces a predicted state transition $\mathcal{S}_{n,h}=\mathcal{T}(\mathcal{S}_{n,h-1},\mathbf{p}_{n,h})$, $h=1,\ldots,H$,  
where $\mathcal{T}$ represents the predicted update of the belief state resulting from observing the target from the corresponding viewpoint. The value of a candidate sequence is defined as
\begin{equation}
J
\left(
\boldsymbol{\Xi}_n\mid\mathcal{S}_n
\right)
=\sum_{h=1}^{H}
\gamma^{h-1}
U
\left(
\mathcal{S}_{n,h-1},\mathbf{p}_{n,h}
\right),
\label{eq:sequence_objective}
\end{equation}
where $U$ denotes the viewpoint utility and $\gamma\in(0,1]$ is a temporal discount factor. The selected sequence is therefore
\begin{equation}
\boldsymbol{\Xi}_n^*
=\arg\max_{\boldsymbol{\Xi}_n}
J
\left(
\boldsymbol{\Xi}_n\mid\mathcal{S}_n
\right).
\label{eq:rh_opt}
\end{equation}

Unlike single-step NBV, the utility of each viewpoint is evaluated using the predicted state resulting from the preceding viewpoints in the sequence. Thus, the value of a viewpoint depends not only on its immediate observation but also on the information expected to have been acquired earlier in the horizon.

Despite optimizing the complete sequence, RH-NBV executes only its first viewpoint, $\mathbf{p}_{n}^{\mathrm{exec}}=\mathbf{p}_{n,1}^*$,

with orientation $\mathbf{q}_{n^{\mathrm{exec}}}=\operatorname{LookAt}(\mathbf{p}_{n^{\mathrm{exec}},\mathbf{t}})$. The resulting real RGB-D observation is fused into the belief state, $\mathcal{S}_{n+1}=\mathcal{T}_{\mathrm{real}}(\mathcal{S}_n,\mathbf{p}_{n}^{\mathrm{exec}})$,
after which the remaining viewpoints are discarded and the optimization is repeated from the updated state until the sensing budget is exhausted or the target information objective is satisfied.

\subsection{Candidate Viewpoint Sequence Generation}
\label{sec:sequence_generation}

We solve Eq.~\eqref{eq:rh_opt} using random-sampling shooting. At each planning iteration, $K$ candidate sequences of horizon $H$ are sampled from the admissible camera-position space $\mathcal{V}_C$ and evaluated using the sequence objective. The highest-utility sequence is selected, although only its first viewpoint is executed.

The first viewpoint of each sequence is sampled globally from $\mathcal{V}_C$. Subsequent viewpoints are generated using a mixture of local tangential motions and global sampling. With probability $b$, a tangential orbit step is sampled approximately perpendicular to the current camera-to-target direction; otherwise, a new position is sampled globally from $\mathcal{V}_C$. The orbit step is given by
\begin{equation}
\mathbf{p}_{h+1}
=\mathbf{p}_{h}
+
\alpha_h
\frac{
\mathbf{d}_h\times\mathbf{u}_h
}{
\left|
\mathbf{d}_h\times\mathbf{u}_h
\right|_2
},
\label{eq:orbit_step}
\end{equation}
where $\mathbf d_h$ is the camera-to-target direction, $\mathbf u_h$ an auxiliary direction, and $\alpha_h$ the step size. Invalid samples are projected onto $\mathcal V_C$. Local orbit steps maintain spatially coherent motion, while global samples enable rapid, larger viewpoint changes between viewing regions to overcome occlusion.

\subsection{Semantic and Occlusion-Aware Viewpoint Utility}
\label{sec:utility}

Candidate sequences generated as described in Sec.~\ref{sec:sequence_generation} are evaluated using a target-centric utility that combines semantic uncertainty with visibility through occluding geometry. The belief map $\mathcal M$ stores occupancy $p_o(i)$ and target-class probability $p_s(i)$; semantic uncertainty is binary Shannon entropy,
\begin{equation}
\mathcal{H}_s(i)
= -p_s(i)\log_2 p_s(i)-(1-p_s(i))\log_2(1-p_s(i)).
\label{eq:semantic_entropy}
\end{equation}

For a candidate position $\mathbf{p}$, rays are cast through the target ROI. The contribution of a sampled point is weighted by its ray transmittance,
\begin{equation}
T(i)=\prod_{j<i}\left(1-p_o(j)\right),
\label{eq:transmittance}
\end{equation}
which represents the probability that the ray reaches $i$ without encountering occupied geometry. The resulting semantic information utility is,
\begin{equation}
U_{\mathrm{sem}}(\mathcal{M},\mathbf{p})
=\frac{1}{|\mathcal{R}|N_r}
\sum_{r\in\mathcal{R}}
\sum_{i=1}^{N_r}
T_{r,i}\mathcal{H}_s(r,i).
\label{eq:semantic_utility}
\end{equation}

To additionally favor viewpoints that can reveal currently occluded target regions, we use an occlusion-aware term $B(\mathcal{M},\mathbf{p})$, which assigns higher utility to regions that are likely occupied but remain semantically uncertain. The final viewpoint utility is
\begin{equation}
U(\mathcal{M},\mathbf{p})
=g\left(U_{\mathrm{sem}}(\mathcal{M},\mathbf{p})\right)
+
w_{\mathrm{occ}}B(\mathcal{M},\mathbf{p}),
\label{eq:total_utility}
\end{equation}
where $g(\cdot)$ denotes the applied normalization and $w_{\mathrm{occ}}$ controls the contribution of the occlusion-aware term.

\subsection{In-Horizon State Prediction}
\label{sec:state_prediction}

During sequence evaluation, RH-NBV predicts the belief-state update associated with each hypothetical viewpoint. Starting from the current map $\mathcal{M}^{(0)}=\mathcal{M}_t$, the predicted state is recursively updated as %
\begin{equation}
\mathcal{M}^{(h)}
=\operatorname{PredictUpdate}
\left(
\mathcal{M}^{(h-1)},\mathbf{p}_h
\right).
\label{eq:predict_update}
\end{equation}
The update approximates the uncertainty reduction expected from a hypothetical observation rather than attempting to reconstruct unseen geometry explicitly. For an updated voxel quantity $x$ (occupancy or semantic probability), the prediction is
\begin{equation}
x^{(h)}
=(1-\eta)x^{(h-1)}
+
\eta\hat{x},
\label{eq:belief_update}
\end{equation}
where $\hat{x}$ is the predicted observation and $\eta$ is the update fraction. Only voxels intersected by the hypothetical observation are updated; all others remain unchanged.

Each candidate sequence maintains its own predicted map. Thus, the utility of a later viewpoint is evaluated after accounting for the information predicted to have been acquired by preceding viewpoints, reducing repeated credit for already observed regions. The predicted updates are used only during sequence evaluation; after execution, the actual RGB-D and semantic observations are fused into the real map.

\subsection{Sequence Evaluation}
\label{sec:sequence_evaluation}
For a candidate sequence $\boldsymbol{\Xi}=[\mathbf p_1,\ldots,\mathbf p_H]$, the predicted state is propagated according to Eq.~\eqref{eq:predict_update}. The sequence is then scored using the utility obtained at each predicted state together with a motion cost,
\begin{equation}
J(\boldsymbol{\Xi}\mid\mathcal M_t)=\sum_{h=1}^{H}
\gamma^{h-1}
\left[
U(\mathcal M^{(h-1)},\mathbf p_h)
-\lambda d_h
\right],
\label{eq:sequence_score}
\end{equation}
where $d_h=\|\mathbf p_h-\mathbf p_{h-1}\|_2$ (with $\mathbf p_0=\mathbf p_t^{\mathrm{current}}$)
The discount factor $\gamma$ reduces the contribution of later hypothetical observations, while $\lambda$ determines the trade-off between expected perceptual utility and camera motion.

The best sequence is selected according to $\boldsymbol{\Xi}_t^*=\arg\max_{\boldsymbol{\Xi}\in\mathcal{X}_H}J(\boldsymbol{\Xi}\mid\mathcal M_t)$, where $\mathcal{X}_H$ denotes the set of feasible sequences generated by the candidate-sampling procedure.

\subsection{Receding-Horizon Execution}
\label{sec:execution}

Although RH-NBV evaluates complete sequences of $H$ future viewpoints, only the first viewpoint of the best sequence is executed. Specifically, $\mathbf p_t^{\mathrm{exec}}=\mathbf p_{t,1}^*$, with camera orientation determined by $\mathbf q_t^{\mathrm{exec}}=\operatorname{LookAt}(\mathbf p_t^{\mathrm{exec}},\mathbf t)$.

\begin{figure}[H]
    \centering
    \begin{subfigure}[b]{0.60\columnwidth}
        \centering
        \includegraphics[
            width=\linewidth,
            trim=5 5 5 5,
            clip
        ]{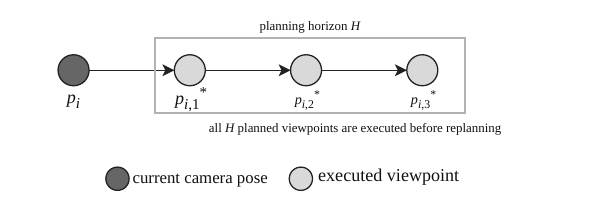}
        \caption{Full-horizon execution.}
        \label{fig:full_horizon}
    \end{subfigure}
    \par\vspace{1mm}
    \begin{subfigure}[b]{0.8\columnwidth}
        \centering
        \includegraphics[
            width=\linewidth,
            trim=30 5 30 5,
            clip
        ]{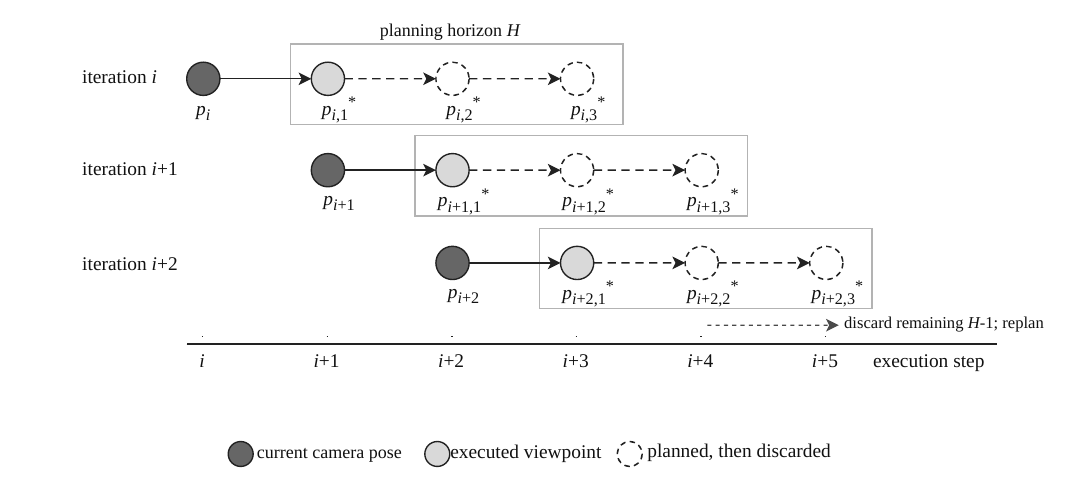}
        \caption{Receding-horizon execution.}
        \label{fig:receding_horizon}
    \end{subfigure}
    \vspace{-1mm}
    \caption{Execution strategies:
    (a) full-horizon execution,
    (b) receding-horizon execution.}
    \label{fig:execution_strategies}
\end{figure}

The manipulator moves the camera to this pose and acquires an RGB-D observation. The resulting observation and semantic prediction are fused into the actual reconstruction state, $\mathcal M_{t+1}=\mathcal T_{\mathrm{real}}(\mathcal M_t,\mathbf p_t^{\mathrm{exec}})$.
The remaining viewpoints are discarded, and new candidate sequences are optimized from the updated state $\mathcal M_{t+1}$. This receding-horizon execution exploits multi-step predictions while continuously correcting them with new observations. Fig.~\ref{fig:execution_strategies} illustrates the receding-horizon strategy (b), in which only the first viewpoint is executed before replanning, unlike the full-sequence execution (a), where all $H$ viewpoints are executed before replanning.

\subsection{Stagnation Escape}
\label{sec:escape}
To prevent local sampling from repeatedly selecting viewpoints within an uninformative region, RH-NBV monitors the improvement in ROI coverage. Let $C_t$ denote the coverage at iteration $t$: if $C_t-C_{t-1}<\tau$ for $P$ consecutive iterations, an escape procedure is triggered. The planner samples $N_{\mathrm{esc}}$ reachable positions satisfying
\begin{equation}
|\mathbf{p}-\mathbf{p}_t|*2\geq d*{\mathrm{esc}},
\end{equation}
and selects the farthest candidate as the new camera position. RH-NBV then resumes planning from the relocated position. Algorithm~\ref{alg:rhnbv} summarizes the complete planning procedure, and Table~\ref{tab:params} reports the fixed hyperparameters.
\begin{algorithm}[t]
\caption{RH-NBV: one planning iteration}
\label{alg:rhnbv}
\scriptsize
\begin{algorithmic}[1]

\Require Current state $\mathcal{S}_n$, current camera position $\mathbf{p}_n$, fixed target point $\mathbf{t}$;
$H,K,\gamma,\lambda,s,b,\mathcal{V}_C$

\State $J_{\mathrm{best}}\gets-\infty,\quad
       \boldsymbol{\Xi}_{n}^{*}\gets\emptyset$

\For{$k=1$ \textbf{to} $K$}

    \State $\boldsymbol{\Xi}_{n}^{(k)}
    \gets
    \Call{GenerateSequence}
    {\mathbf{p}_n,H,\mathcal{V}_C,s,b}$

    \State $\mathcal{S}_{n,0}^{(k)}
    \gets\Call{Copy}{\mathcal{S}_n},\quad
    J^{(k)}\gets0,\quad
    \mathbf{p}_{n,0}^{(k)}\gets\mathbf{p}_n$

    \For{$h=1$ \textbf{to} $H$}

        \State $\mathbf{p}_{n,h}^{(k)}
        \gets\boldsymbol{\Xi}_{n}^{(k)}[h]$

        \State $\mathbf{q}_{n,h}^{(k)}
        \gets\Call{LookAt}
        {\mathbf{p}_{n,h}^{(k)},\mathbf{t}}$

        \State $J^{(k)} \gets J^{(k)}
        +\gamma^{h-1}
        \Bigl(
        U(\mathcal{S}_{n,h-1}^{(k)},\mathbf{p}_{n,h}^{(k)})
        -\lambda
        \|\mathbf{p}_{n,h}^{(k)}
        -\mathbf{p}_{n,h-1}^{(k)}\|_2
        \Bigr)$

        \State $\mathcal{S}_{n,h}^{(k)}
        \gets
        \Call{PredictUpdate}
        {\mathcal{S}_{n,h-1}^{(k)},\mathbf{p}_{n,h}^{(k)}}$

    \EndFor

    \If{$J^{(k)}>J_{\mathrm{best}}$}
        \State $J_{\mathrm{best}}\gets J^{(k)},\quad
        \boldsymbol{\Xi}_{n}^{*}\gets\boldsymbol{\Xi}_{n}^{(k)}$
    \EndIf

\EndFor

\State $\mathbf{p}_{n}^{\mathrm{exec}}
\gets\boldsymbol{\Xi}_{n}^{*}[1],\quad
\mathbf{q}_{n}^{\mathrm{exec}}
\gets\Call{LookAt}{\mathbf{p}_{n}^{\mathrm{exec}},\mathbf{t}}$

\State \Return
$(\mathbf{p}_{n}^{\mathrm{exec}},
\mathbf{q}_{n}^{\mathrm{exec}})$

\end{algorithmic}
\end{algorithm}

\begin{table}[t]
\centering
\caption{Default RH-NBV hyperparameters.}
\label{tab:params}
\begin{tabular}{@{}llc@{}}
\toprule
Symbol & Meaning & Value \\
\midrule
$H$ & Number of future viewpoints in each candidate sequence & 3 \\
$K$ & Number of candidate viewpoint sequences per iteration & 10 \\
$\gamma$ & Discount factor for future viewpoint utilities & 0.85 \\
$\lambda$ & Weight of the camera-motion cost & 2.0 \\
$w_{\mathrm{occ}}$ & Weight of the occlusion-aware utility term & 2.0 \\
$s$ & Maximum tangential orbit-step length (m) & 0.065 \\
$b$ & Probability of generating a tangential orbit step & 0.7 \\
$P$ & Number of low-improvement iterations before escape & 4 \\
$\tau$ & Minimum ROI-coverage improvement per iteration (\%) & 1.5 \\
\bottomrule
\end{tabular}
\end{table}

\section{Experimental Setup}
\label{sec:setup}

\subsection{Simulation and Physical Setup}
\label{sec:simulation_setup}

\begin{figure}[t]
\centering
\includegraphics[width=\linewidth]{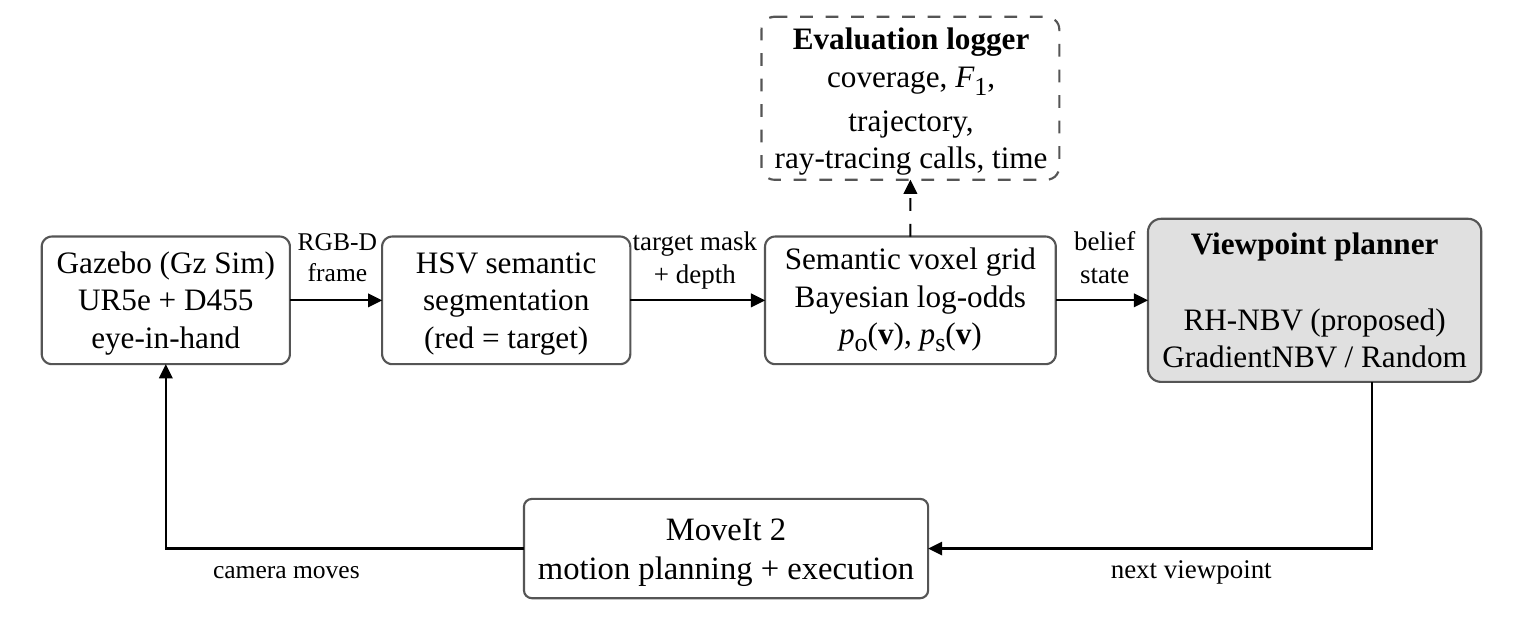}
\caption{Perception--planning--action loop of the proposed RH-NBV planner.}
\label{fig:pipeline}
\end{figure}

\begin{figure}[t]
\centering
\includegraphics[width=0.8\linewidth]{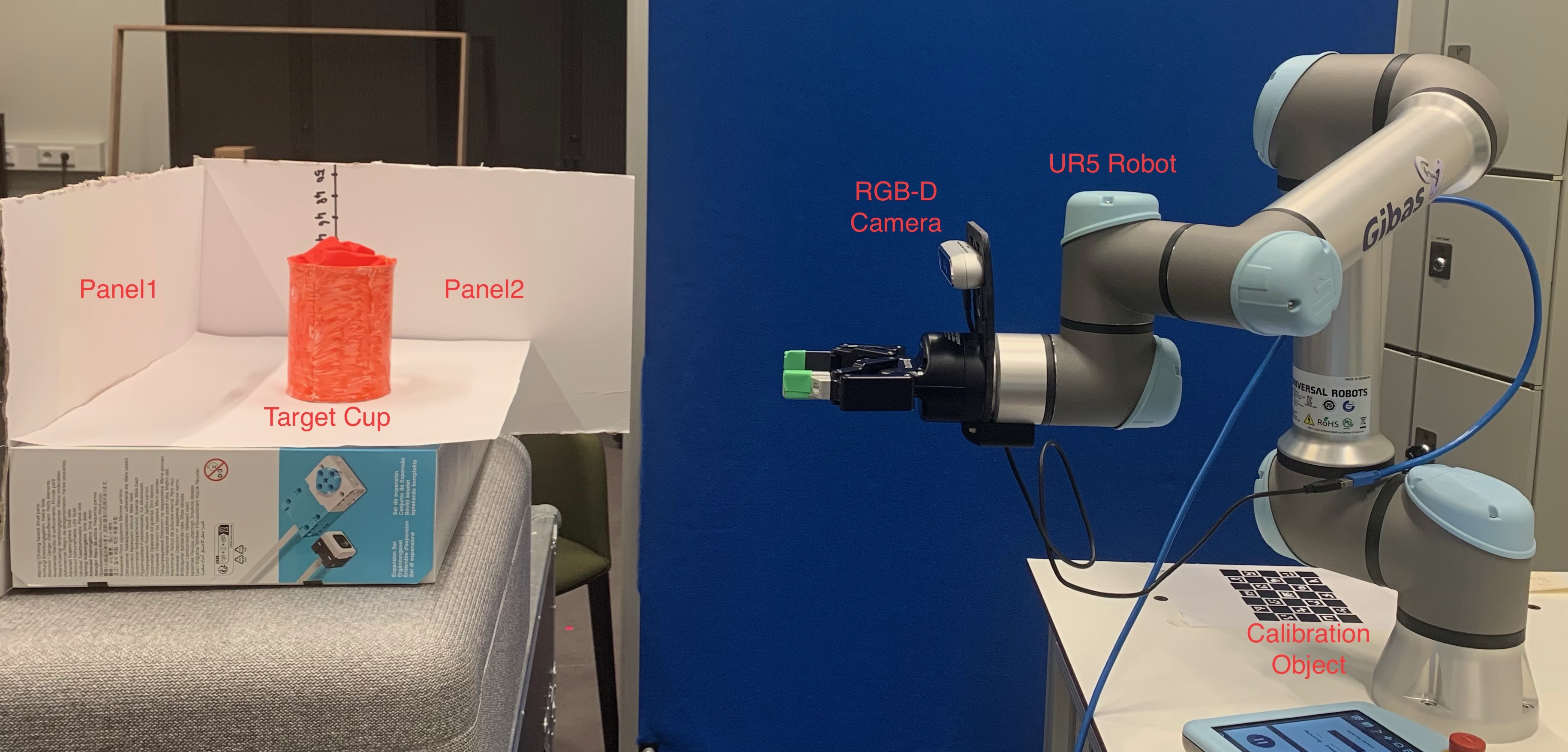}
\caption{Physical experimental setup with the UR5e manipulator and eye-in-hand RealSense D455 camera.}
\label{fig:physical_setup}
\end{figure}

\begin{figure}[h]
\centering
\begin{subfigure}[b]{0.32\linewidth}\centering
\includegraphics[width=\linewidth,height=1.9cm,keepaspectratio]{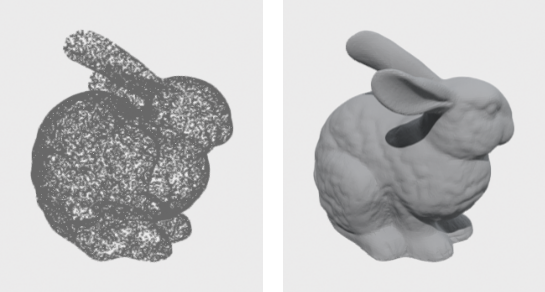}
\caption{Bunny}
\end{subfigure}\hfill 
\begin{subfigure}[b]{0.32\linewidth}\centering
\includegraphics[width=\linewidth,height=1.9cm,keepaspectratio]{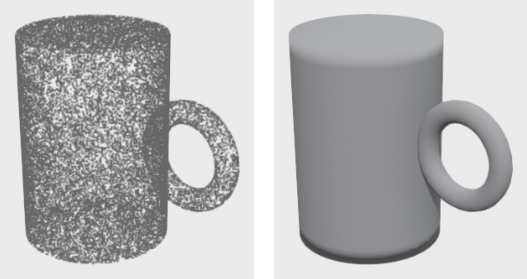}
\caption{Coffee mug}
\end{subfigure}\hfill 
\begin{subfigure}[b]{0.32\linewidth}\centering
\includegraphics[width=\linewidth,height=1.9cm,keepaspectratio]{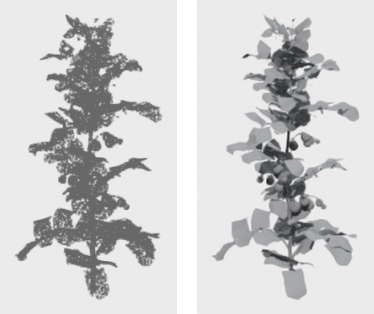}
\caption{Tomato plant}
\end{subfigure}
\caption{The three reconstruction targets used in our experiments. For each object, the point cloud sampled from its surface mesh (left) and the 3D model as spawned in the Gazebo simulation (right) are shown.}
\label{fig:targets}
\end{figure}

RH-NBV is implemented in ROS~2 Jazzy and evaluated in both simulation and on a physical UR5e manipulator equipped with an eye-in-hand Intel RealSense D455 RGB-D camera. The target ROI is extracted in 3D by combining HSV-based RGB segmentation with aligned depth data. RH-NBV evaluates candidate viewpoints using a voxel belief map of the workspace, while MoveIt~2 executes the selected camera motions. The simulation environment is implemented in Gazebo, while the physical setup uses the same planning and perception pipeline with a real camera and robot. Simulation uses a Stanford Bunny and coffee mug for controlled object-level occlusion and a tomato plant for natural self-occlusion as shown in Fig.~\ref{fig:targets}. Physical experiments are performed with the coffee mug to validate RH-NBV on a real robot and RGB-D sensor. Fig.~\ref{fig:pipeline} summarizes the perception--planning--action loop, and Fig.~\ref{fig:physical_setup} shows the physical experimental setup.

\subsection{Evaluation Scenarios}
\label{sec:evaluation}

\begin{figure}[h]
    \centering
    \begin{minipage}[t]{0.245\columnwidth}
        \centering\includegraphics[width=\linewidth,height=\linewidth]
        {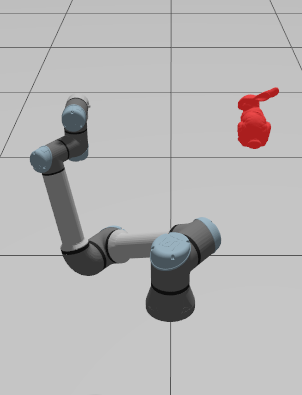}\\[-1mm]
        {\scriptsize (a) Unoccluded}
    \end{minipage}\hfill
    \begin{minipage}[t]{0.245\columnwidth}
        \centering
\includegraphics[width=\linewidth,height=\linewidth]
        {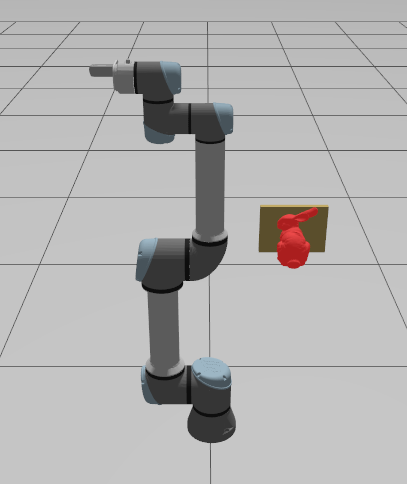}\\[-1mm]
        {\scriptsize (b) Right-side wall}
    \end{minipage}\hfill
    \begin{minipage}[t]{0.245\columnwidth}
        \centering
        \includegraphics[width=\linewidth,height=\linewidth]
        {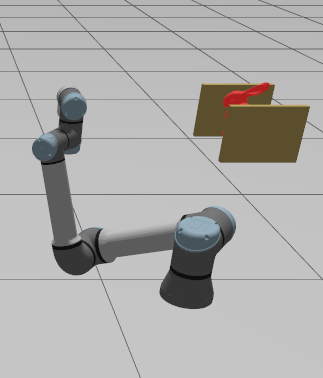}\\[-1mm]
        {\scriptsize (c) Tunnel}
    \end{minipage}\hfill
    \begin{minipage}[t]{0.245\columnwidth}
        \centering
        \includegraphics[width=\linewidth,height=\linewidth]
        {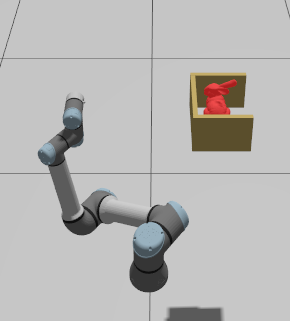}\\[-1mm]
        {\scriptsize (d) U-shape}
    \end{minipage}

    \vspace{2mm}

    \begin{minipage}[t]{0.245\columnwidth}
        \centering
        \includegraphics[width=\linewidth,height=\linewidth]
        {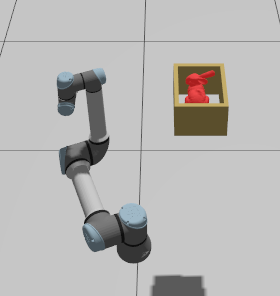}\\[-1mm]
        {\scriptsize (e) Four walls}
    \end{minipage}\hfill
    \begin{minipage}[t]{0.245\columnwidth}
        \centering
        \includegraphics[width=\linewidth,height=\linewidth]
        {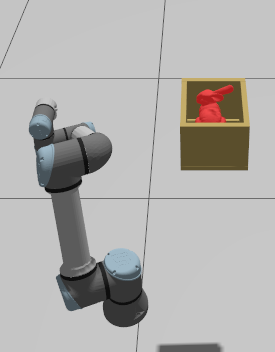}\\[-1mm]
        {\scriptsize (f) Bottom lid}
    \end{minipage}\hfill
    \begin{minipage}[t]{0.245\columnwidth}
        \centering
        \includegraphics[width=\linewidth,height=\linewidth]
        {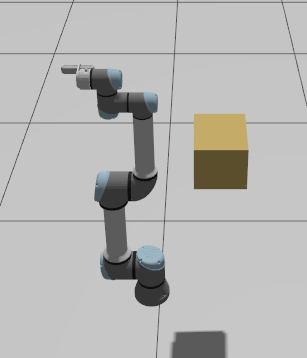}\\[-1mm]
        {\scriptsize (g) Fully enclosed}
    \end{minipage}\hfill
    \begin{minipage}[t]{0.245\columnwidth}
        \centering
        \includegraphics[width=\linewidth,height=\linewidth]
        {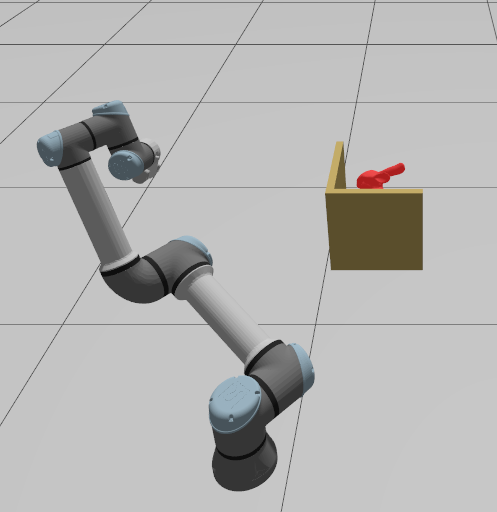}\\[-1mm]
        {\scriptsize (h) L-shape}
    \end{minipage}

    \caption{Simulated object-reconstruction scenarios in Gazebo using a UR5 arm equipped with an eye-in-hand RGB-D camera. The configurations progress from the unoccluded scene in (a) to increasingly occluded scenes in (b)--(g), where each successive stage retains the previous occluders and adds an additional wall, resulting in progressively reduced target visibility. Configuration (h) shows an L-shaped occluder with a different spatial arrangement.}
    \label{fig:scenarios}
\end{figure}

\textbf{Controlled external occlusion.}
The Stanford Bunny and coffee mug are evaluated in simulation using a six-stage panel ladder. The first stage contains the unobstructed target, while subsequent stages progressively add panels and reduce the directly visible target surface. An additional L-shaped occluder tests a more irregular occlusion geometry. Fig.~\ref{fig:scenarios} illustrates the staged bunny configurations. This progression evaluates whether the advantage of non-myopic planning increases as target visibility becomes more constrained.

\textbf{Natural self-occlusion.}
The tomato plant is evaluated in simulation without artificial occluders. Four plant orientations and three target levels---whole plant, blossoms, and fruits---are considered. The target level defines the semantic ROI, while the plant structure provides the occlusion. This scenario evaluates the behavior of RH-NBV under naturally occurring self-occlusion and varying semantic targets.

\subsection{Comparison Methods}
\label{sec:comparison}

RH-NBV is compared with three alternative viewpoint-selection strategies. \textbf{GradientNBV}~\cite{burusa_henten_gert_kootstra_2024} is a single-step gradient-ascent planner that optimizes the same semantic, occlusion-aware utility as RH-NBV, while \textbf{Random} selects a nearby admissible viewpoint without information-gain evaluation. All methods used the same voxel representation, target ROI, admissible camera workspace, viewpoint definition, initial pose, perception and mapping pipeline, and execution budget of $N=20$ viewpoints, with Trial~0 using a fixed starting position. The comparison therefore isolates the planning strategy: GradientNBV makes myopic single-view decisions, whereas RH-NBV evaluates multi-view sequences using predicted state updates. Random provides a utility-free reference.
\subsection{Evaluation Metrics}
\label{sec:metrics}
We evaluate target information recovery, reconstruction quality, and motion efficiency using four primary metrics. \textbf{ROI coverage} is the fraction of target-ROI voxels observed by the executed viewpoints. \textbf{$F_1$ reconstruction score} quantifies geometric agreement between the reconstructed target and reference geometry. \textbf{Occluded-target recall} is the fraction of target surface points initially hidden from the camera that are subsequently observed. \textbf{Trajectory distance} is the cumulative Euclidean distance traveled by the camera.
As secondary metrics, we report \textbf{motion-success rate}, the fraction of commanded camera motions successfully executed by \textit{MoveIt~2}; and \textbf{ray-tracing calls}, which quantify the computational effort of viewpoint evaluation.

\subsection{Experimental Protocol}
\label{sec:protocol}

For each target and occlusion configuration, all planners start from the same initial pose and use the same sensing budget, target ROI, and semantic target definition. RGB-D observations are fused into a common reconstruction pipeline before each subsequent viewpoint is selected. RH-NBV executes only the first viewpoint of each selected sequence and replans from the updated state; the remaining viewpoints are discarded.

The experiments address three questions: (i) whether multi-step planning improves target information recovery over myopic viewpoint selection, (ii) whether this advantage increases with occlusion severity, and (iii) whether the improvement arises from the planning horizon rather than increased computational effort. Controlled occlusion, natural self-occlusion, and horizon-ablation experiments are used to evaluate these effects. Default planner parameters are reported in Table~\ref{tab:params}.


\section{Results}
\label{sec:results}

We evaluate the planners from three perspectives: performance under controlled external occlusion, the effect of planning horizon, and robustness to natural self-occlusion.

\subsection{Controlled Occlusion}
\label{sec:controlled_results}
Table~\ref{tab:mug-bunny-burusa} summarizes the results for the coffee mug and Stanford bunny. RH-NBV consistently achieves higher ROI coverage, $F_1$, and occluded-target recall than GradientNBV and Random as the number of executed viewpoints increases. The advantage is particularly pronounced for the more heavily occluded configurations. 
\begin{figure}[h]
    \centering
    \begin{minipage}[t]{0.33\columnwidth}
        \centering
        \includegraphics[width=\linewidth,height=\linewidth]{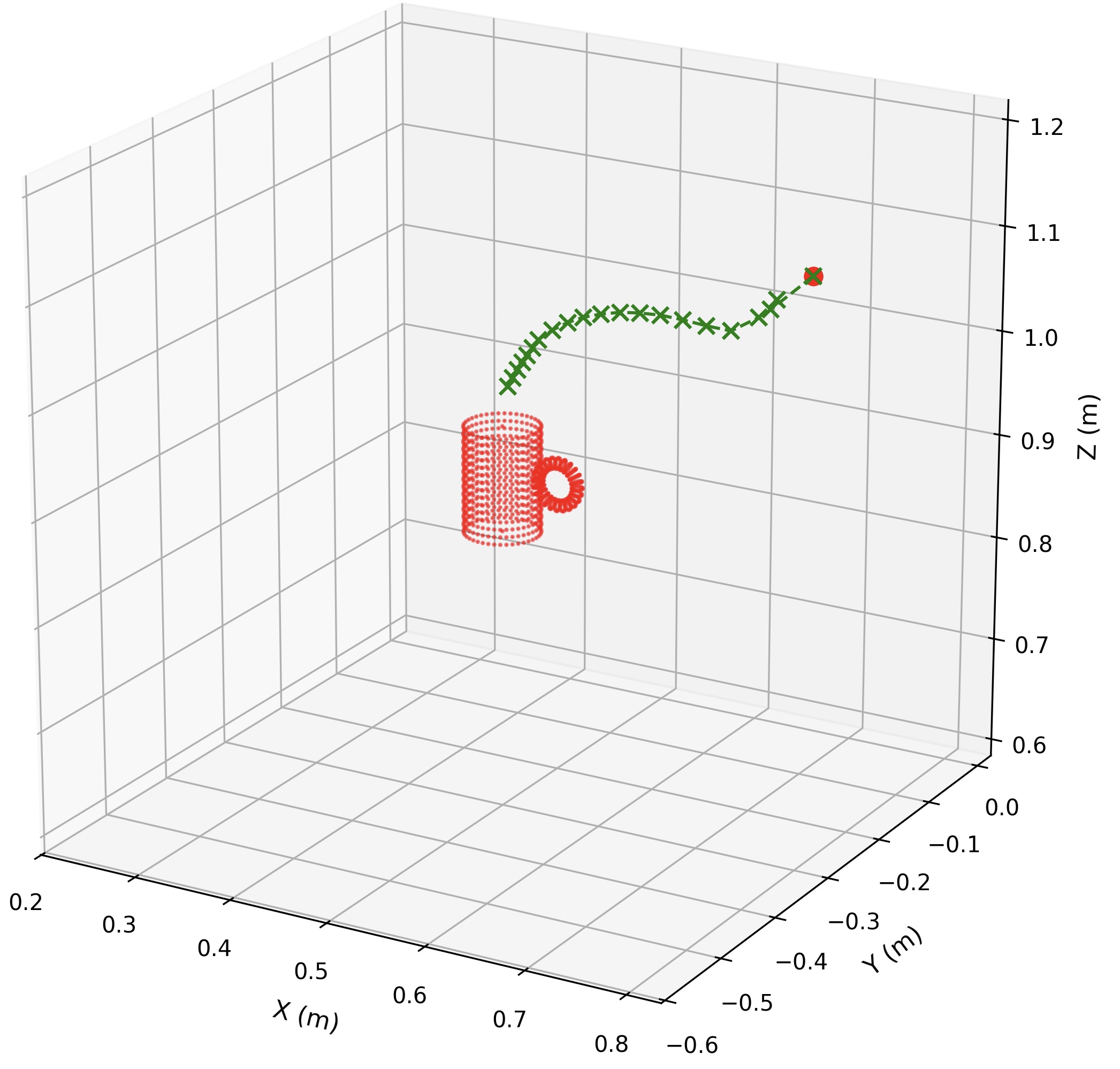}\\[-1mm]
        {\scriptsize (a) GradientNBV trajectory}
\includegraphics[width=\linewidth,height=\linewidth]{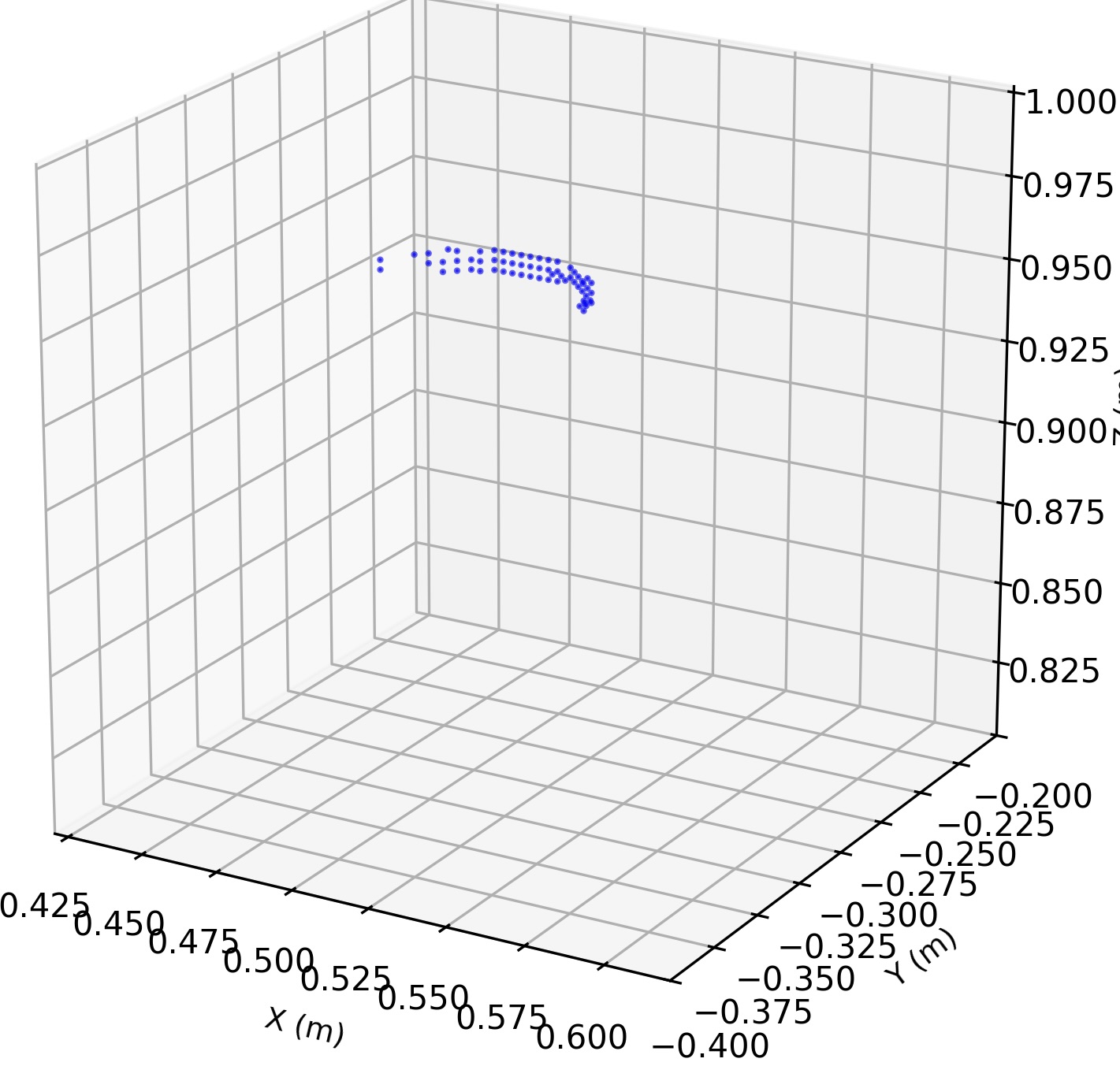}\\[-1mm]
        {\scriptsize (d) GradientNBV reconstruction (64 voxel)}
    \end{minipage}\hfill
    \begin{minipage}[t]{0.33\columnwidth}
        \centering
        \includegraphics[width=\linewidth,height=\linewidth]{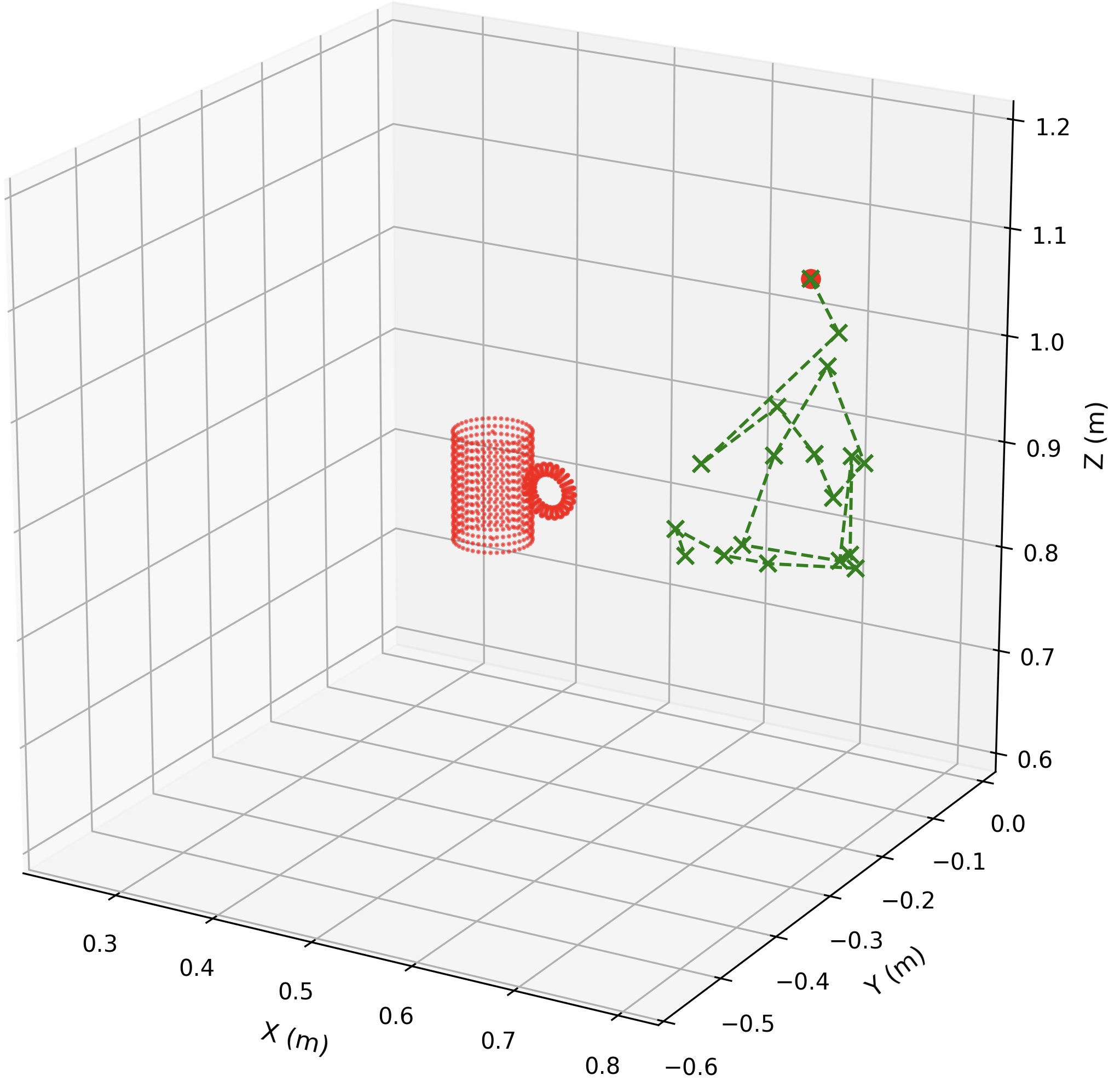}\\[-1mm]
        {\scriptsize (b) Random trajectory}
        \includegraphics[width=\linewidth,height=\linewidth]{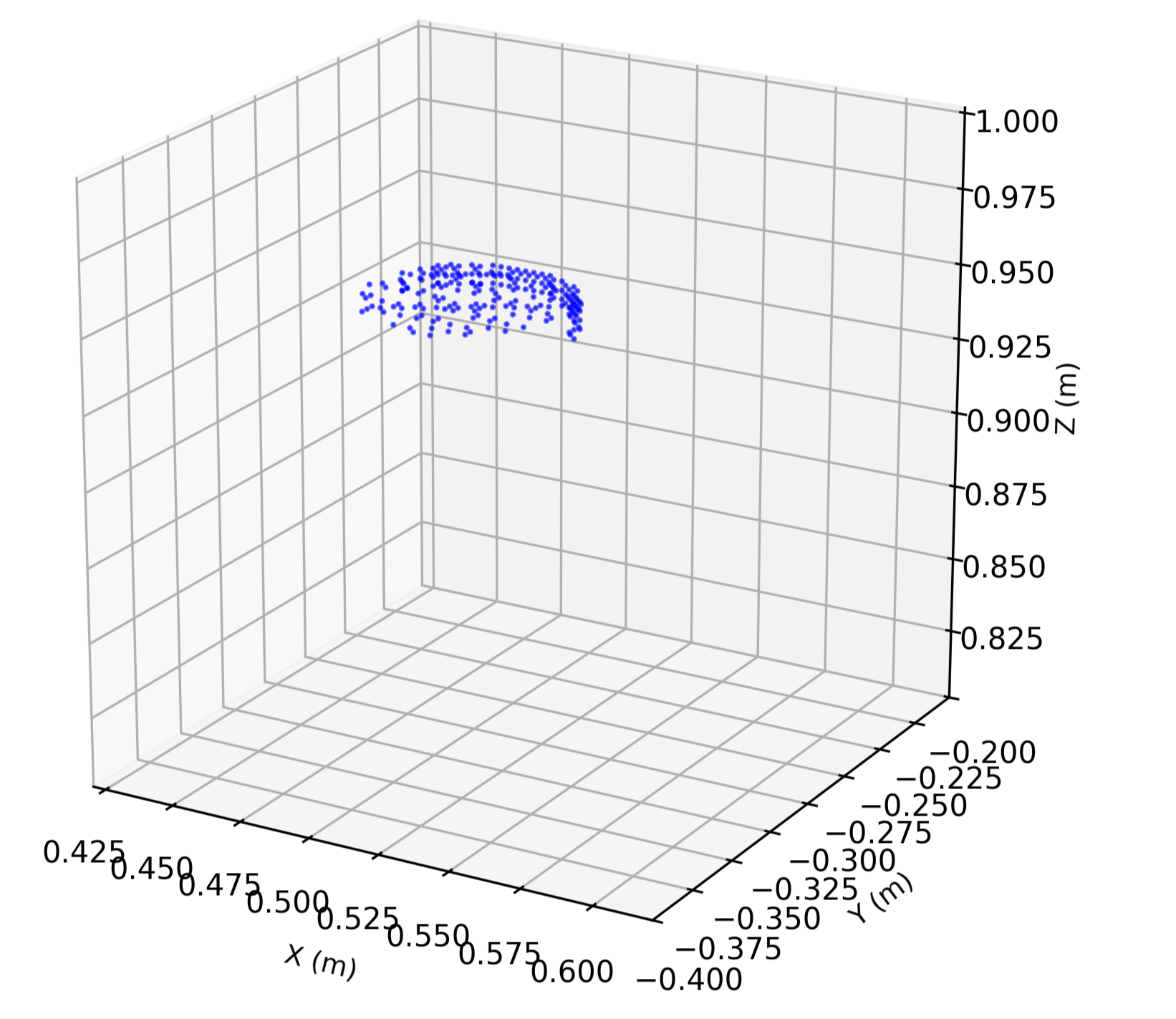}\\[-1mm]
        {\scriptsize (e) Random reconstruction (220 voxel)}
    \end{minipage}\hfill
        \begin{minipage}[t]{0.33\columnwidth}
        \centering
        \includegraphics[width=\linewidth,height=\linewidth]{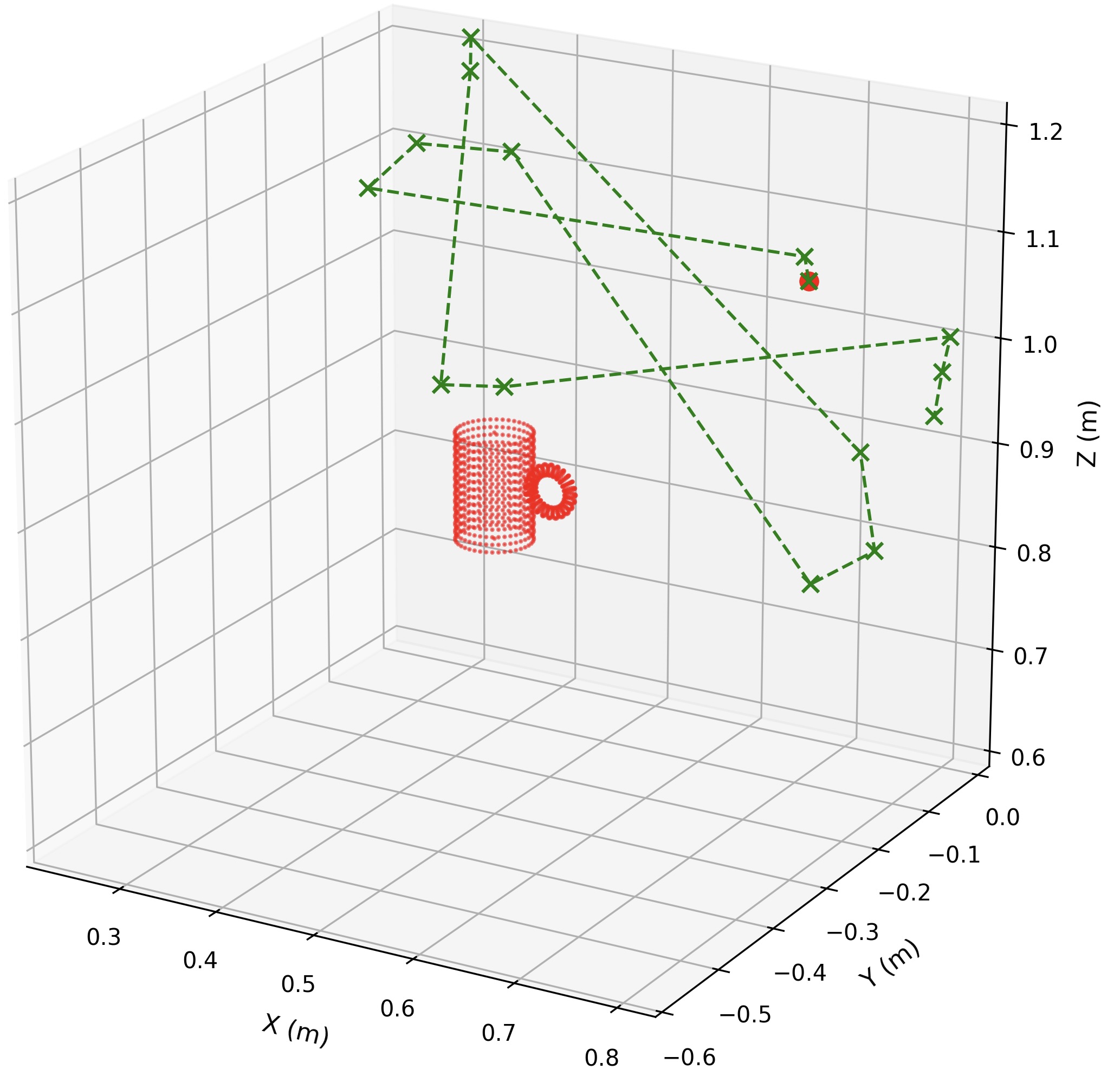}\\[-1mm]
        {\scriptsize (c) RH-NBV trajectory} 
        \includegraphics[width=\linewidth,height=\linewidth]{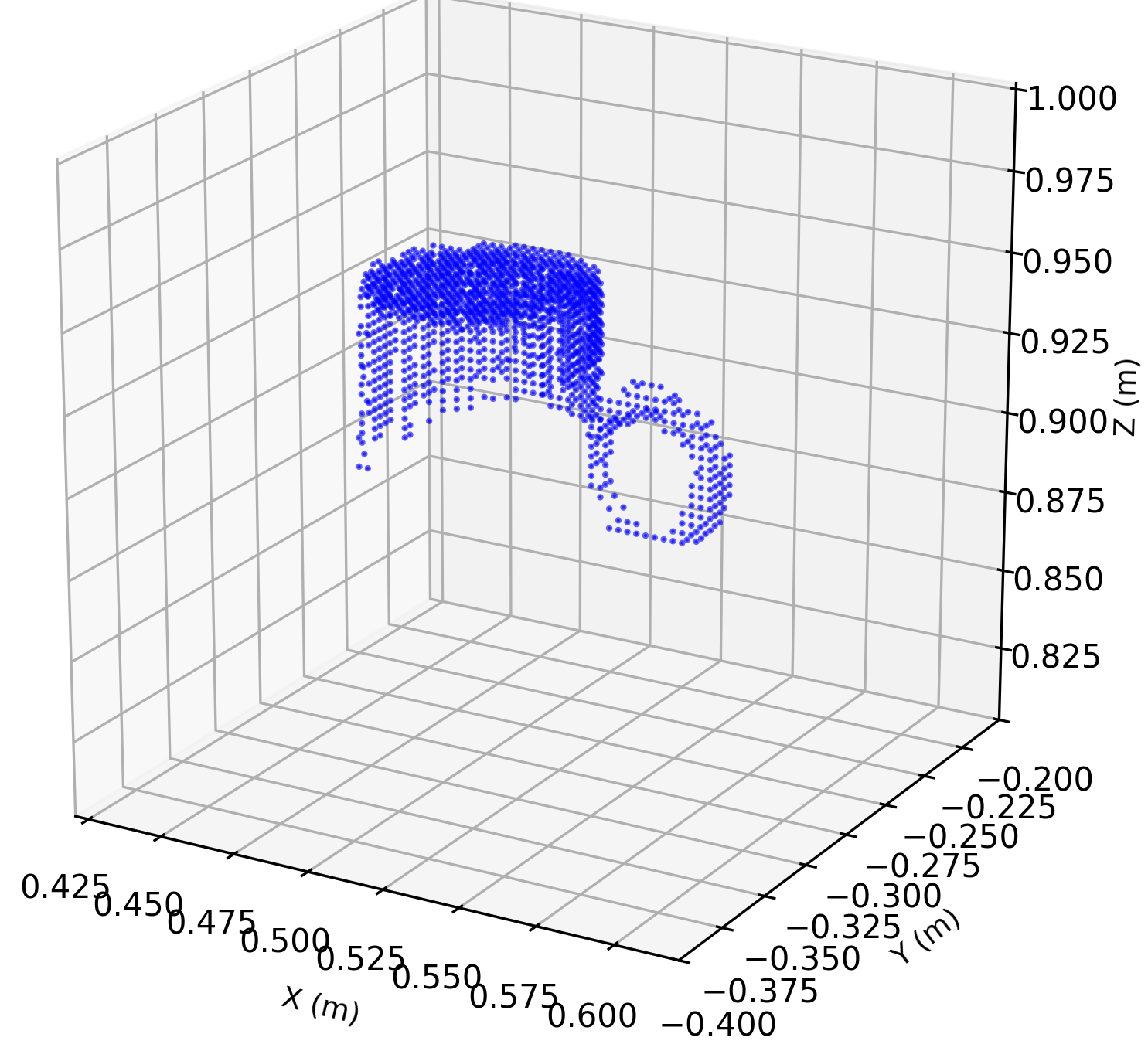}\\[-1mm]
        {\scriptsize (f) RH-NBV reconstruction (1{,}604) voxel}
    \end{minipage}\hfill
    \caption{Coffee-mug reconstruction after 20 viewpoints in the panels5 configuration (Fig. \ref{fig:scenarios}(f)). (a)--(c) show the ground-truth cup with the executed trajectories of GradientNBV, Random, and RH-NBV, respectively, while (d)--(f) show the corresponding reconstructions obtained by GradientNBV, Random, and RH-NBV.}
\label{fig:recon}
\end{figure} 
The results in Table~\ref{tab:ladder} further show that the benefit of RH-NBV depends on occlusion severity. For the coffee mug, the advantage is small under \texttt{panels1} and \texttt{panels2}, but increases substantially under \texttt{panels3}, \texttt{panels5}, and the L-shaped occluder. For example, under \texttt{panels3}, RH-NBV improves coverage from $34.23\%$ to $63.82\%$ and occluded-target recall from $7.4\%$ to 
\begin{table*}[]
\centering
\caption{Performance of RH-NBV and baseline planners for coffee-mug and Stanford-bunny reconstruction. Results are averaged over 80 experiments per target (160 experiments per planner), with performance reported after 0, 5, 10, 15, and 20 executed viewpoints. View 0 denotes the common predefined initial viewpoint. $\uparrow$ indicates higher is better and $\downarrow$ indicates lower is better.}
\label{tab:mug-bunny-burusa}
\resizebox{\textwidth}{!}{%
\begin{tabular}{ll|ccccc|ccccc|ccccc|ccccc|ccccc}
\toprule
& & \multicolumn{5}{c|}{ROI coverage (\%) $\uparrow$}
& \multicolumn{5}{c|}{$F_1$ score (\%) $\uparrow$}
& \multicolumn{5}{c|}{Occluded recall (\%) $\uparrow$}
& \multicolumn{5}{c|}{Ray-tracing calls (\#) $\downarrow$}
& \multicolumn{5}{c}{Trajectory distance (m) $\downarrow$} \\
\textbf{Target} & \textbf{Planner}
& $0$ & $5$ & $10$ & $15$ & $20$
& $0$ & $5$ & $10$ & $15$ & $20$
& $0$ & $5$ & $10$ & $15$ & $20$
& $0$ & $5$ & $10$ & $15$ & $20$
& $0$ & $5$ & $10$ & $15$ & $20$ \\
\midrule
\multirow{4}{*}{Mug}
& RH-NBV
& 0.0 & \textbf{67.1} & \textbf{71.5} & \textbf{75.4} & \textbf{76.0}
& 0.0 & \textbf{81.0} & \textbf{81.8} & \textbf{82.7} & \textbf{83.3}
& 0.0 & \textbf{72.8} & \textbf{74.3} & \textbf{75.7} & \textbf{76.3}
& 0 & 150 & 300 & 450 & 600
& 0.00 & 0.68 & 0.95 & 1.75 & 2.15 \\
& GradientNBV
& 0.0 & 62.3 & 66.6 & 68.9 & 69.4
& 0.0 & 56.7 & 64.8 & 68.0 & 68.1
& 0.0 & 47.8 & 54.6 & 58.6 & 58.8
& 0 & 5 & 10 & 15 & 20
& 0.00 & \textbf{0.18} & \textbf{0.29} & \textbf{0.42} & \textbf{0.50} \\
& Random
& 0.0 & 46.5 & 56.0 & 57.9 & 59.1
& 0.0 & 53.7 & 53.5 & 54.0 & 54.3
& 0.0 & 49.1 & 50.2 & 50.3 & 50.7
& 0 & 0 & 0 & 0 & 0
& 0.00 & 0.39 & 0.80 & 1.30 & 1.62 \\
\midrule
\multirow{4}{*}{Bunny}
& RH-NBV
& 0.0 & \textbf{28.6} & \textbf{34.2} & \textbf{38.9} & \textbf{41.1}
& 0.0 & \textbf{42.8} & \textbf{47.4} & \textbf{50.4} & \textbf{52.8}
& 0.0 & \textbf{33.8} & \textbf{36.9} & \textbf{40.0} & \textbf{42.2}
& 0 & 150 & 300 & 450 & 600
& 0.00 & 0.46 & 0.81 & 1.25 & 1.50 \\
& GradientNBV
& 0.0 & 22.5 & 24.8 & 26.1 & 26.4
& 0.0 & 30.7 & 32.3 & 34.0 & 34.4
& 0.0 & 24.7 & 26.9 & 28.6 & 29.2
& 0 & 5 & 10 & 15 & 20
& 0.00 & \textbf{0.16} & \textbf{0.31} & \textbf{0.37} & \textbf{0.42} \\
& Random
& 0.0 & 21.8 & 30.1 & 32.2 & 33.0
& 0.0 & 32.7 & 43.5 & 43.8 & 43.9
& 0.0 & 24.4 & 34.3 & 34.8 & 35.0
& 0 & 0 & 0 & 0 & 0
& 0.00 & 0.43 & 0.86 & 1.24 & 1.52 \\
\bottomrule
\end{tabular}%
}
\end{table*}
\begin{figure}[h]
    \centering
    \begin{minipage}[t]{0.33\columnwidth}
        \centering
        \includegraphics[width=\linewidth,height=\linewidth]
        {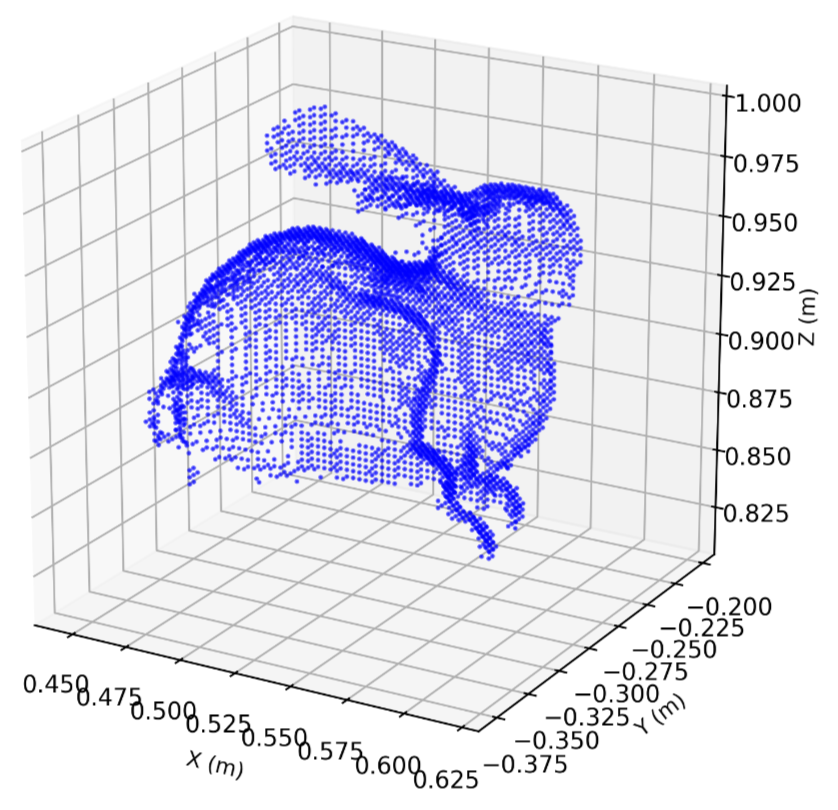}\\[-1mm]\hspace{-3mm}
        {\scriptsize (a) RH-NBV: 5{,}061 voxels}
\includegraphics[width=\linewidth,height=\linewidth]
        {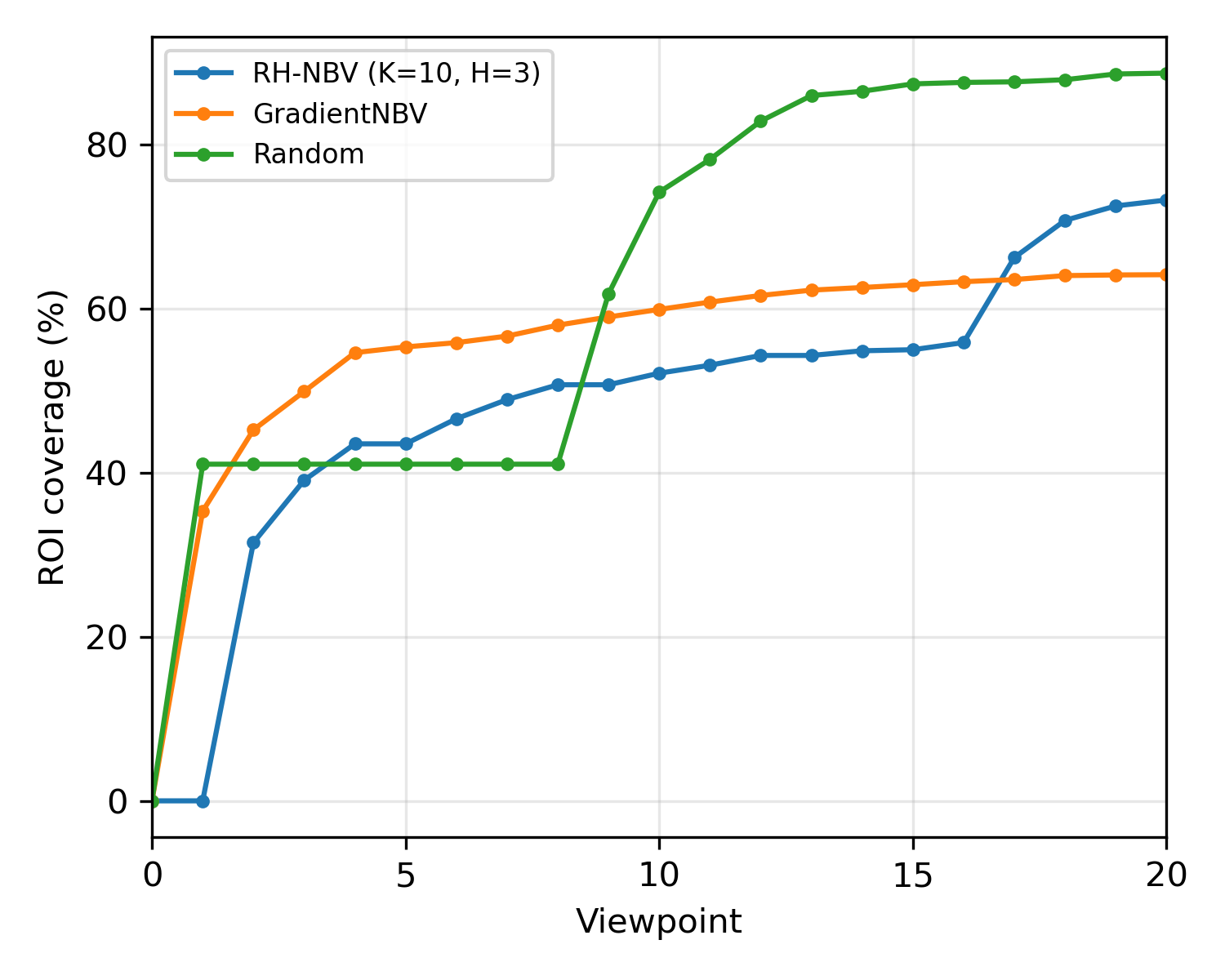}\\[-1mm]
        {\scriptsize (d) ROI coverage}
    \end{minipage}\hfill
    \begin{minipage}[t]{0.33\columnwidth}
        \centering
        \includegraphics[width=\linewidth,height=\linewidth]
        {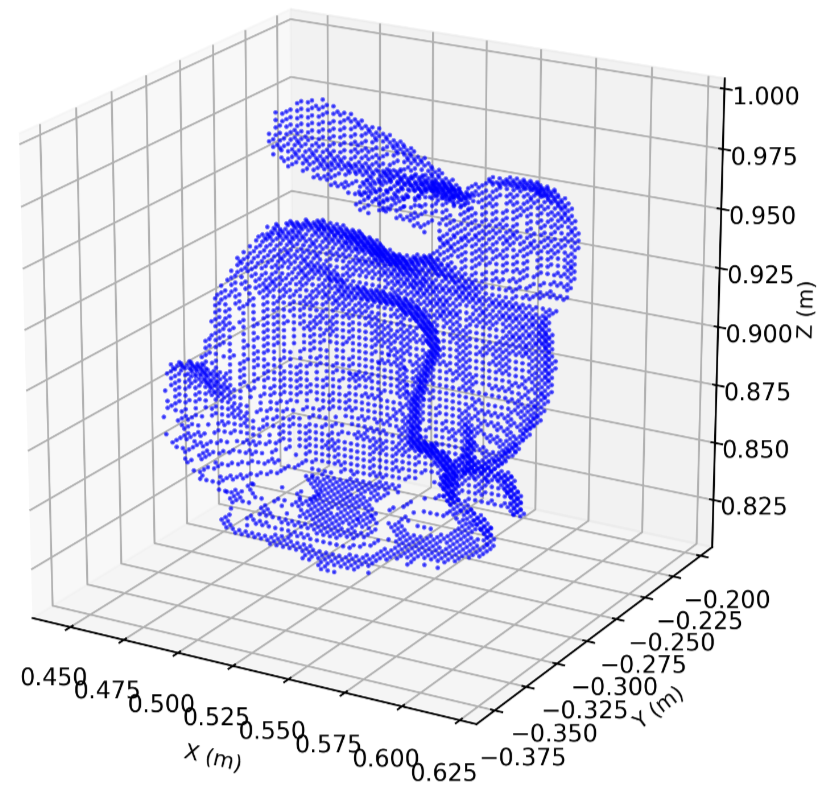}\\[-1mm] \hspace{-4mm}
        {\scriptsize (b) GradientNBV: 5{,}148 voxels}
        \includegraphics[width=\linewidth,height=\linewidth]
        {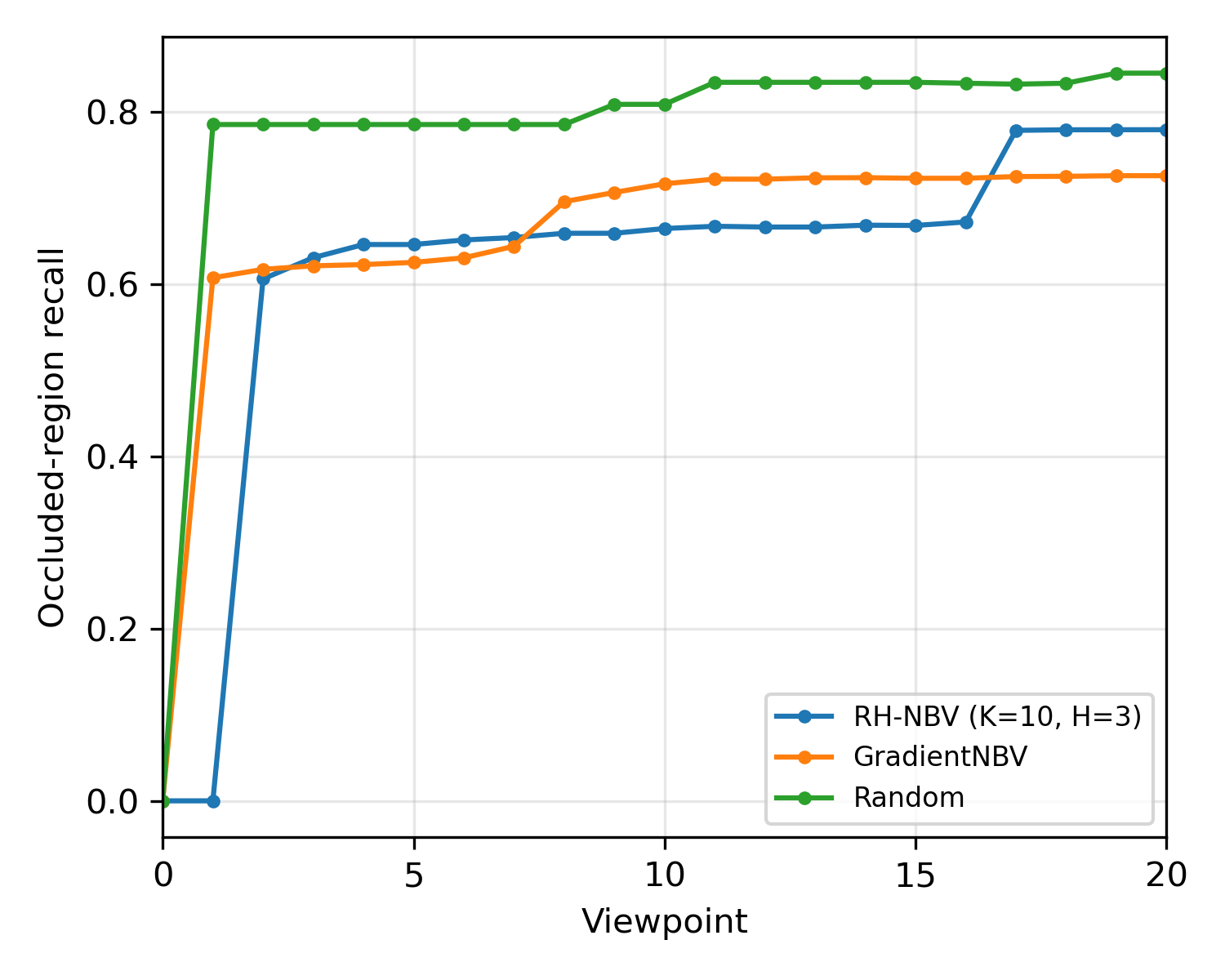}\\[-1mm]
        {\scriptsize (e) Occluded recall}
    \end{minipage}\hfill
        \begin{minipage}[t]{0.33\columnwidth}
        \centering
        \includegraphics[width=\linewidth,height=\linewidth]
        {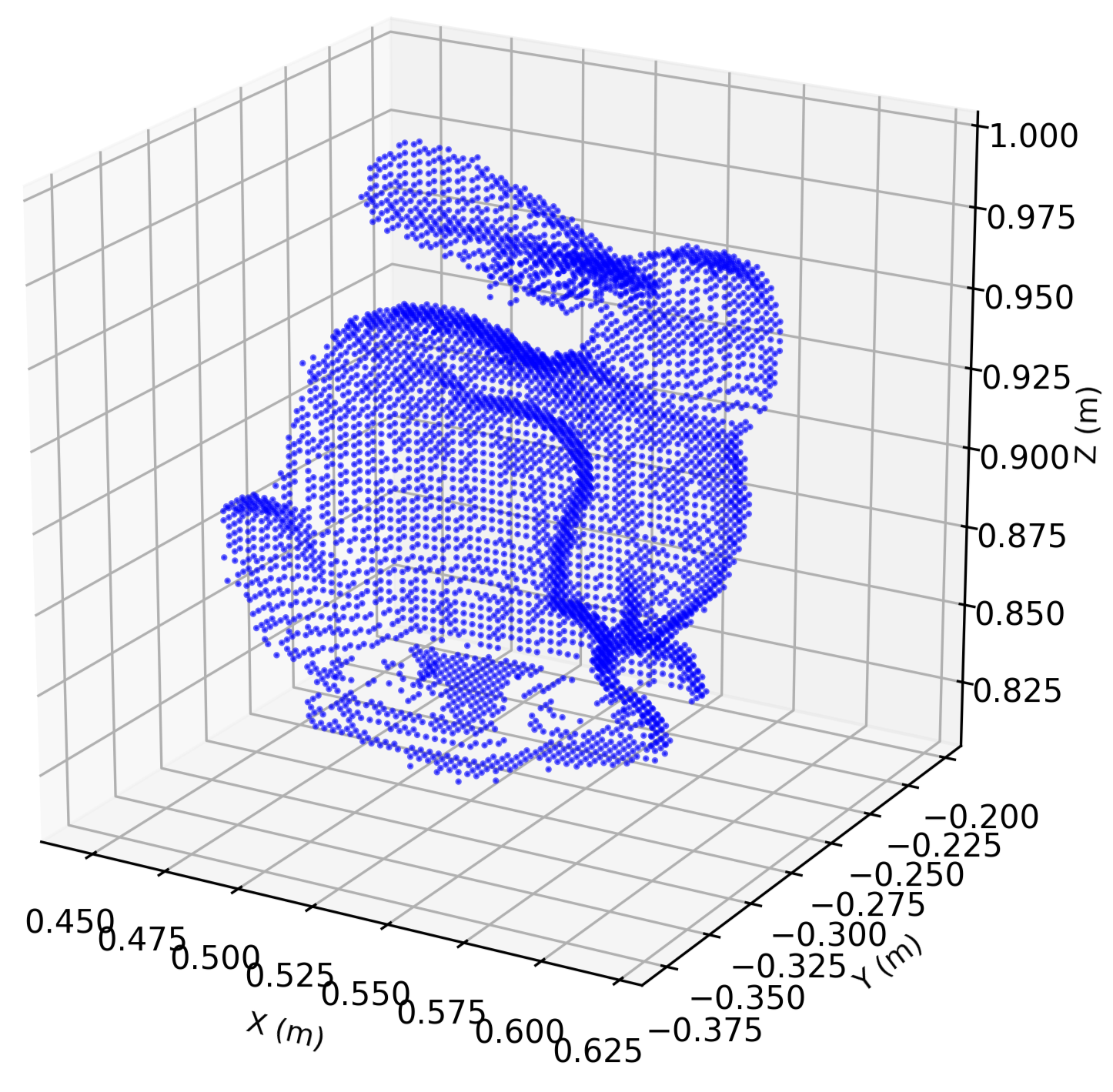}\\[-1mm]
        {\scriptsize (c) Random: 5{,}254 voxels}
        \includegraphics[width=\linewidth,height=\linewidth]
        {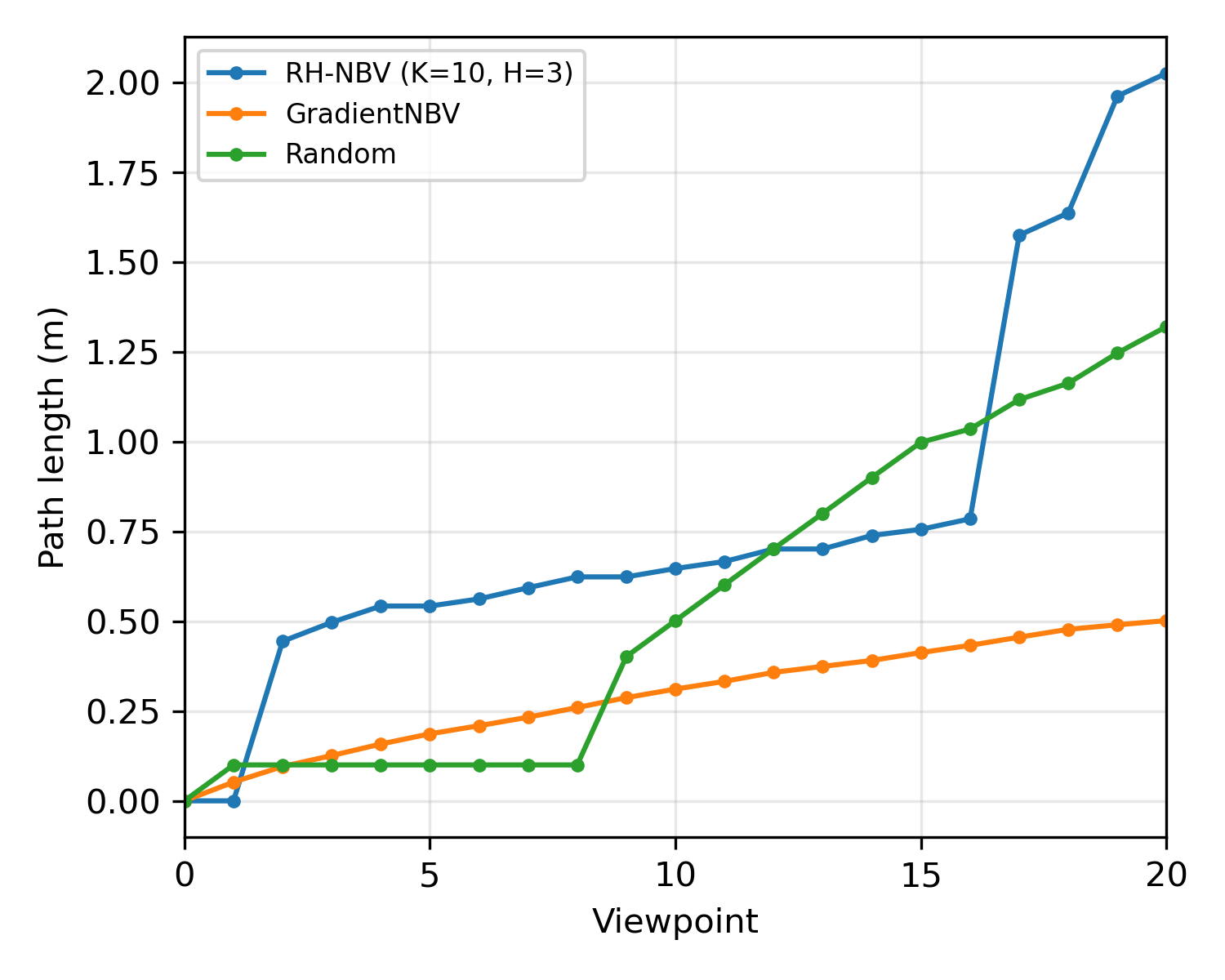}\\[-1mm]
        {\scriptsize (f) Path length}
    \end{minipage}\hfill
    \caption{Easy occlusion scenario (\texttt{panels1}):  Comparison of RH-NBV, GradientNBV, and Random for bunny reconstruction under the panels1 occlusion scenario (Fig. \ref{fig:scenarios}(b)) after 20 viewpoints. Final reconstructions are shown in (a)-(c); per-viewpoint ROI coverage, occluded recall, and trajectory length in (d)-(f). Reconstructions are similar across planners, but the curves reveal differences in visibility gain and motion efficiency. 
       }
\label{fig:panels1}
\end{figure}
\begin{figure}[t]
    \centering
 \begin{minipage}[t]{0.33\columnwidth}
        \centering
        \includegraphics[width=\linewidth,height=\linewidth]{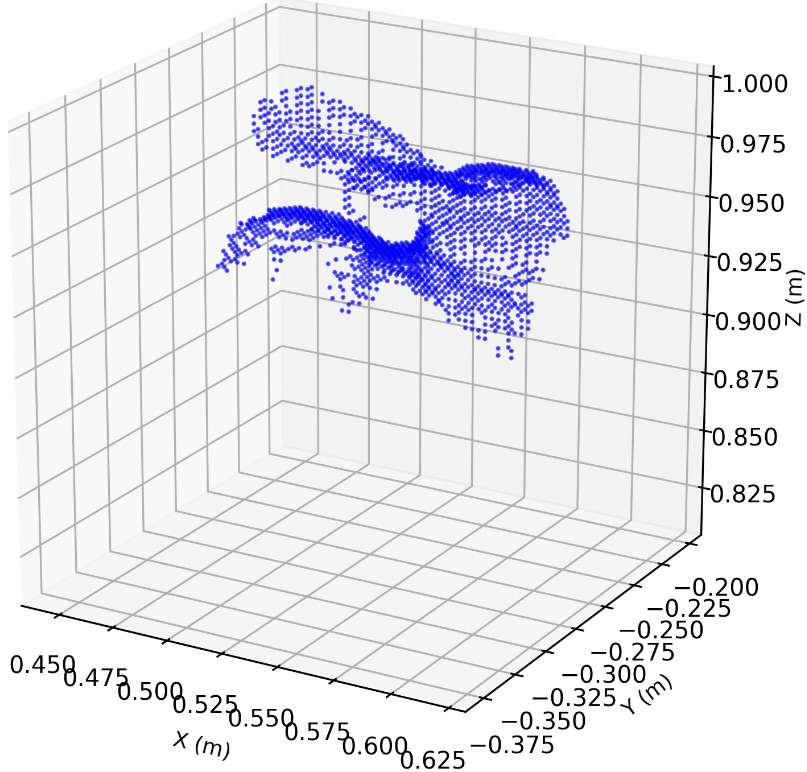}\\[-1mm]
        {\scriptsize (a) RH-NBV: 2{,}156 voxels}
\includegraphics[width=\linewidth,height=\linewidth]
        {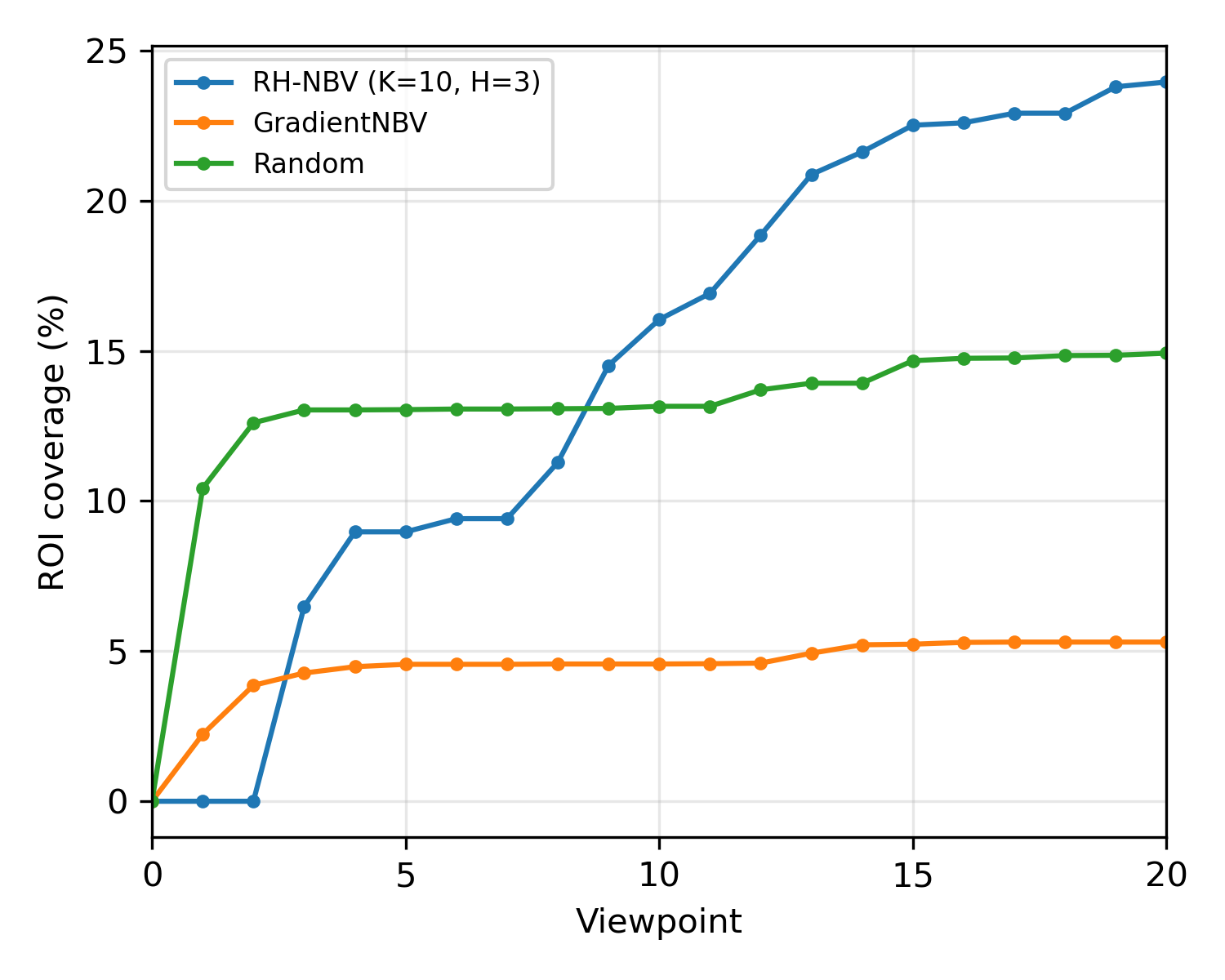}\\[-1mm]
        {\scriptsize (d) ROI coverage}
    \end{minipage}\hfill
     \begin{minipage}[t]{0.33\columnwidth}
        \centering
        \includegraphics[width=\linewidth,height=\linewidth]{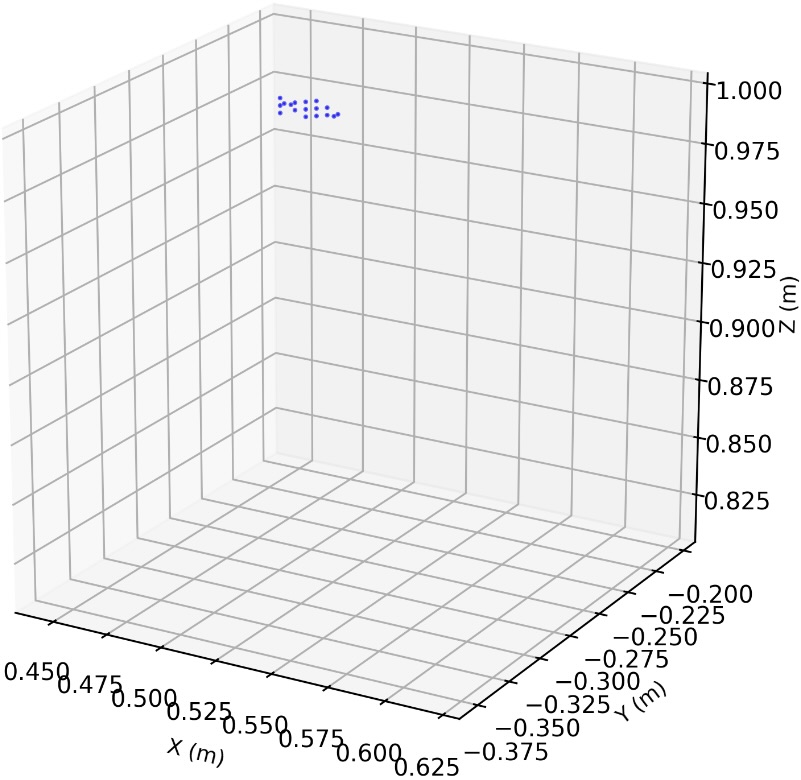}
\\[-1mm]
        {\scriptsize (b) GradientNBV: 17 voxels}
\includegraphics[width=\linewidth,height=\linewidth]
        {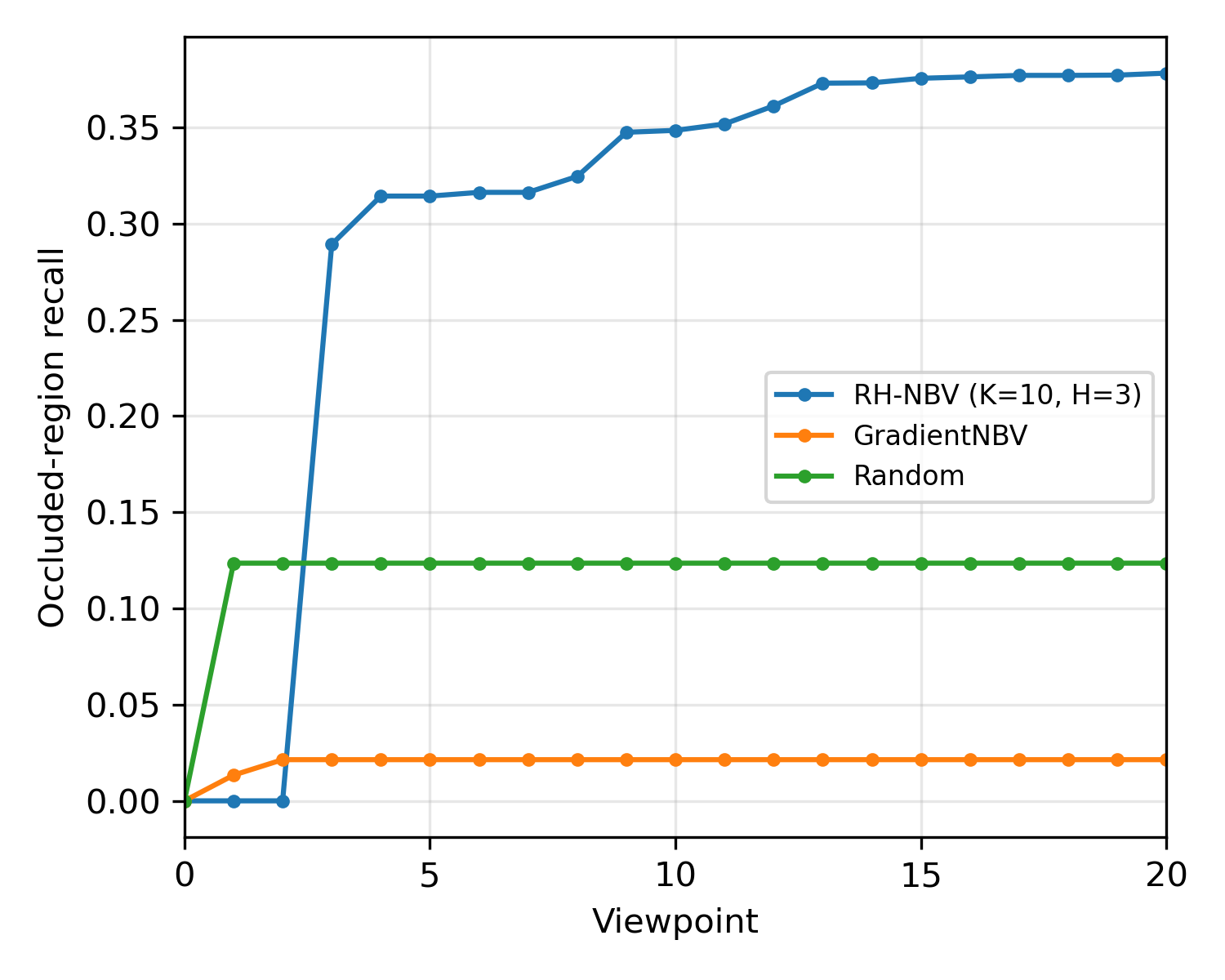}\\[-1mm]
        {\scriptsize (e) Occluded recall}
    \end{minipage}\hfill
     \begin{minipage}[t]{0.33\columnwidth}
        \centering
        \includegraphics[width=\linewidth,height=\linewidth]{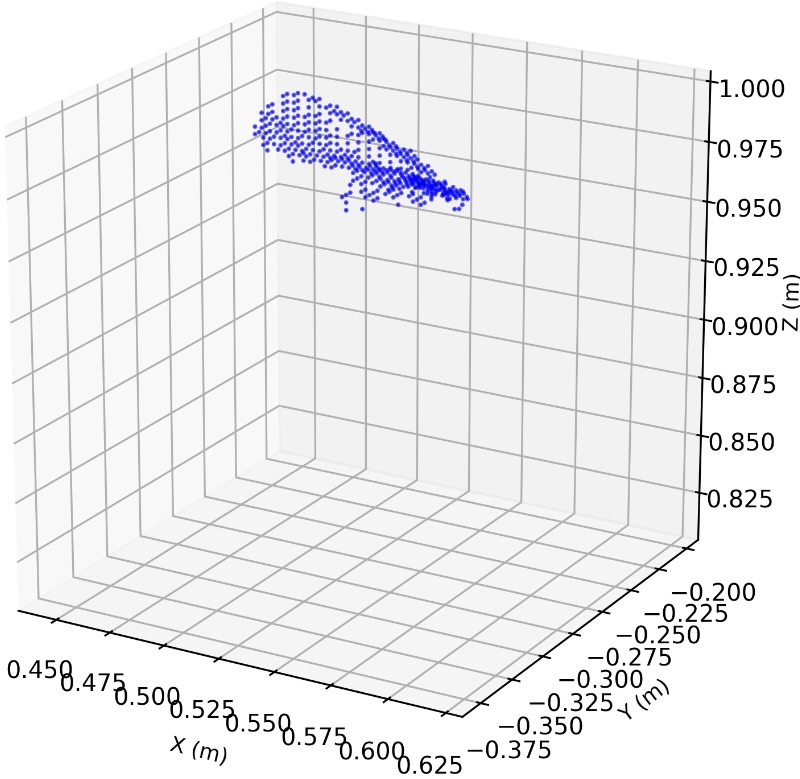}\\[-1mm]
        {\scriptsize (c) Random: 435 voxels}
\includegraphics[width=\linewidth,height=\linewidth]{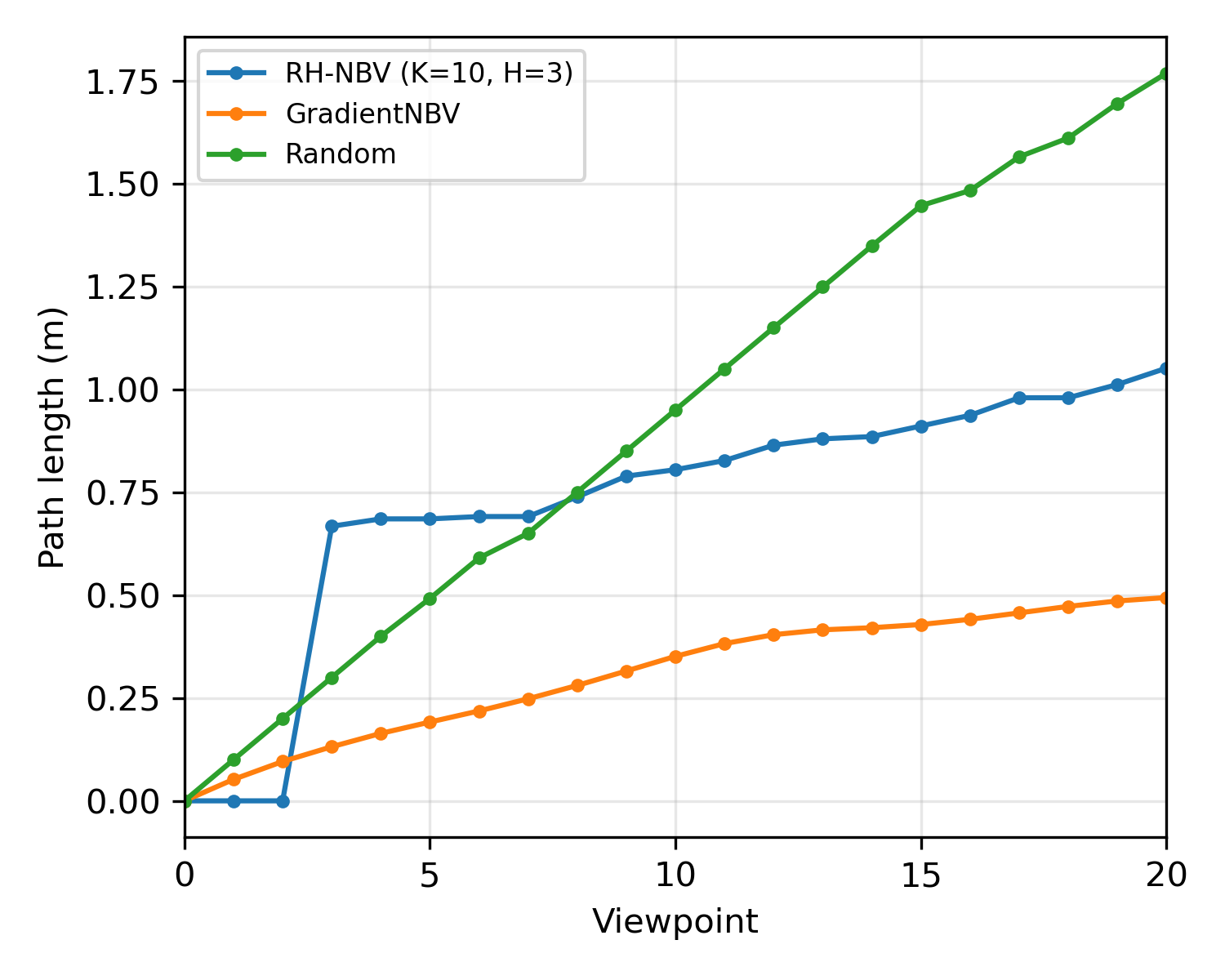}\\[-1mm]
        {\scriptsize (f) Path length.}
    \end{minipage}\hfill
    \caption{Difficult occlusion scenario (\texttt{L-shaped panels}): Reconstruction performance under the challenging L-shaped occluder for the 35{,}947-voxel ground truth bunny after 20 executed viewpoints. Final reconstructions are shown in (a)-(c); per-viewpoint ROI coverage, occluded-region recall, and trajectory length in (d)-(f).
    RH-NBV progressively improves target visibility and reconstruction of the occluded region, whereas the myopic planners show limited improvement after the initial viewpoints.
    }
    \label{fig:difficult}
\end{figure}
$85.5\%$ relative to GradientNBV. The bunny exhibits the same overall trend, although GradientNBV remains competitive under mild occlusion. These results indicate that RH-NBV does not provide a uniform advantage over myopic planning. Its benefit emerges when occlusion makes the value of a viewpoint dependent on observations obtained from preceding viewpoints, supporting multi-step planning. Fig.~\ref{fig:recon} provides a qualitative comparison for the heavily occluded mug, where RH-NBV recovers substantially more of the target surface than the baselines. Figs.~\ref{fig:panels1} and~\ref{fig:difficult} show similar behavior for mild and severe bunny occlusion: the planners are comparable under mild occlusion, while RH-NBV achieves greater visibility and occluded-region recovery under the L-shaped occluder.
\subsection{Natural Self-Occlusion}
\label{sec:natural_occlusion}
\begin{figure*}
\centering
 \begin{minipage}[h]{0.24\columnwidth}
        \centering
\includegraphics[width=\linewidth,height=\linewidth]{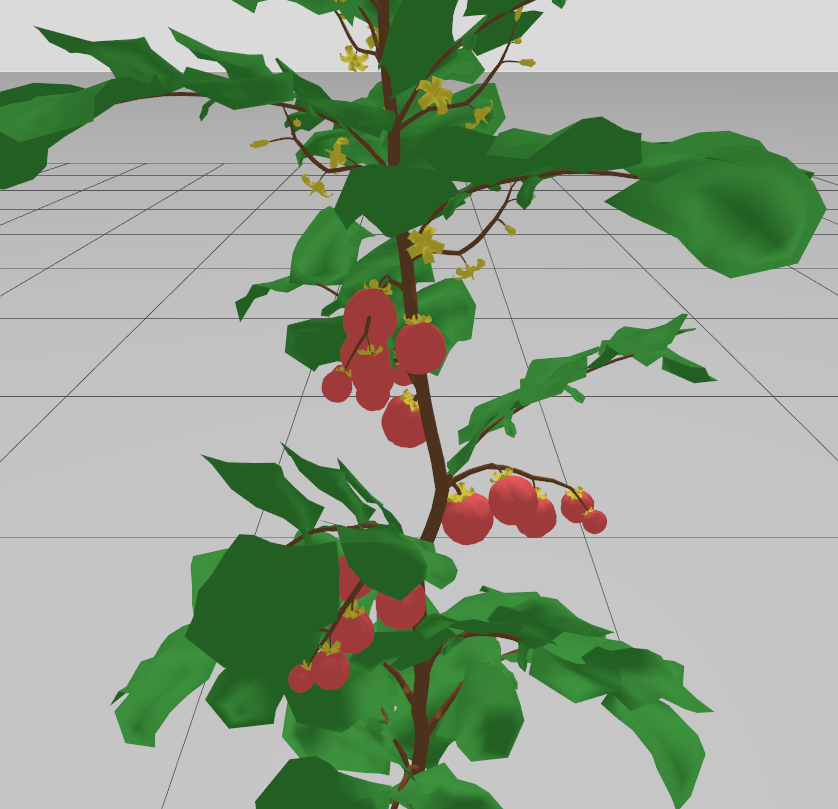}\\[-1mm]
        {\scriptsize  Fruits.}
\includegraphics[width=\linewidth,height=\linewidth]
        {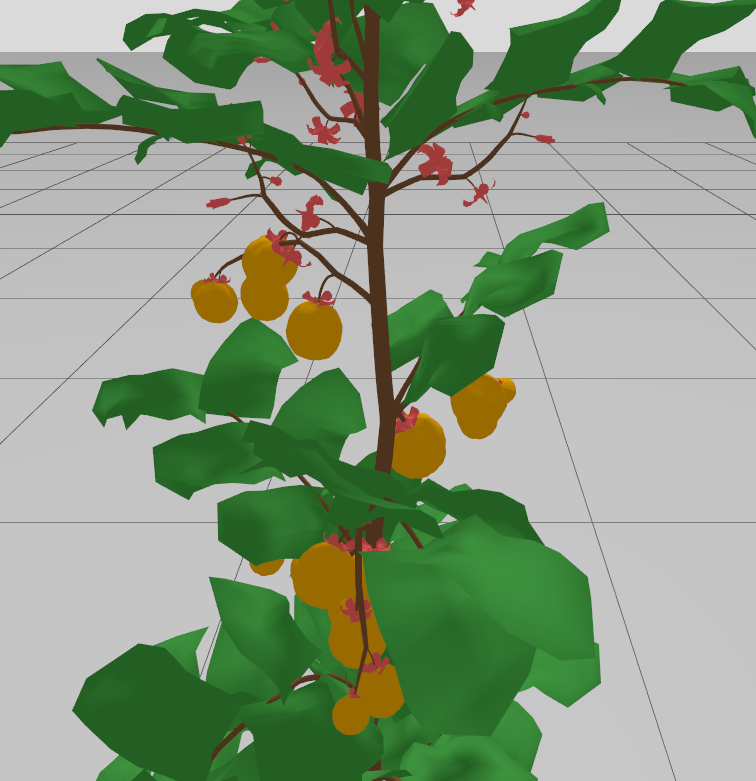}\\[-1mm]
        {\scriptsize Blossoms.}
\includegraphics[width=\linewidth,height=\linewidth]
        {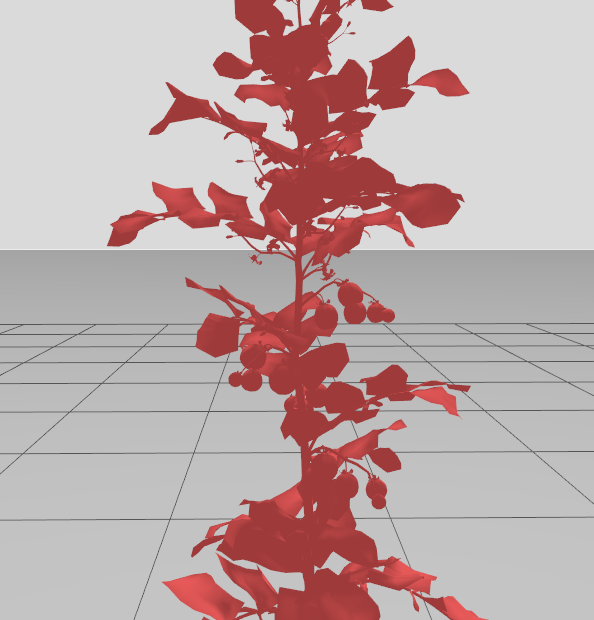}\\[-1mm]
        {\scriptsize Whole Plant.}
    \end{minipage}\hfill
     \begin{minipage}[h]{0.02\columnwidth}
        \centering 
            $\Rightarrow$
            \vspace{10mm}\\
            $\Rightarrow$
            \vspace{10mm}\\
            $\Rightarrow$
    \end{minipage}\hfill
     \begin{minipage}[h]{0.25\columnwidth}
        \centering
 \includegraphics[width=\linewidth,height=\linewidth]{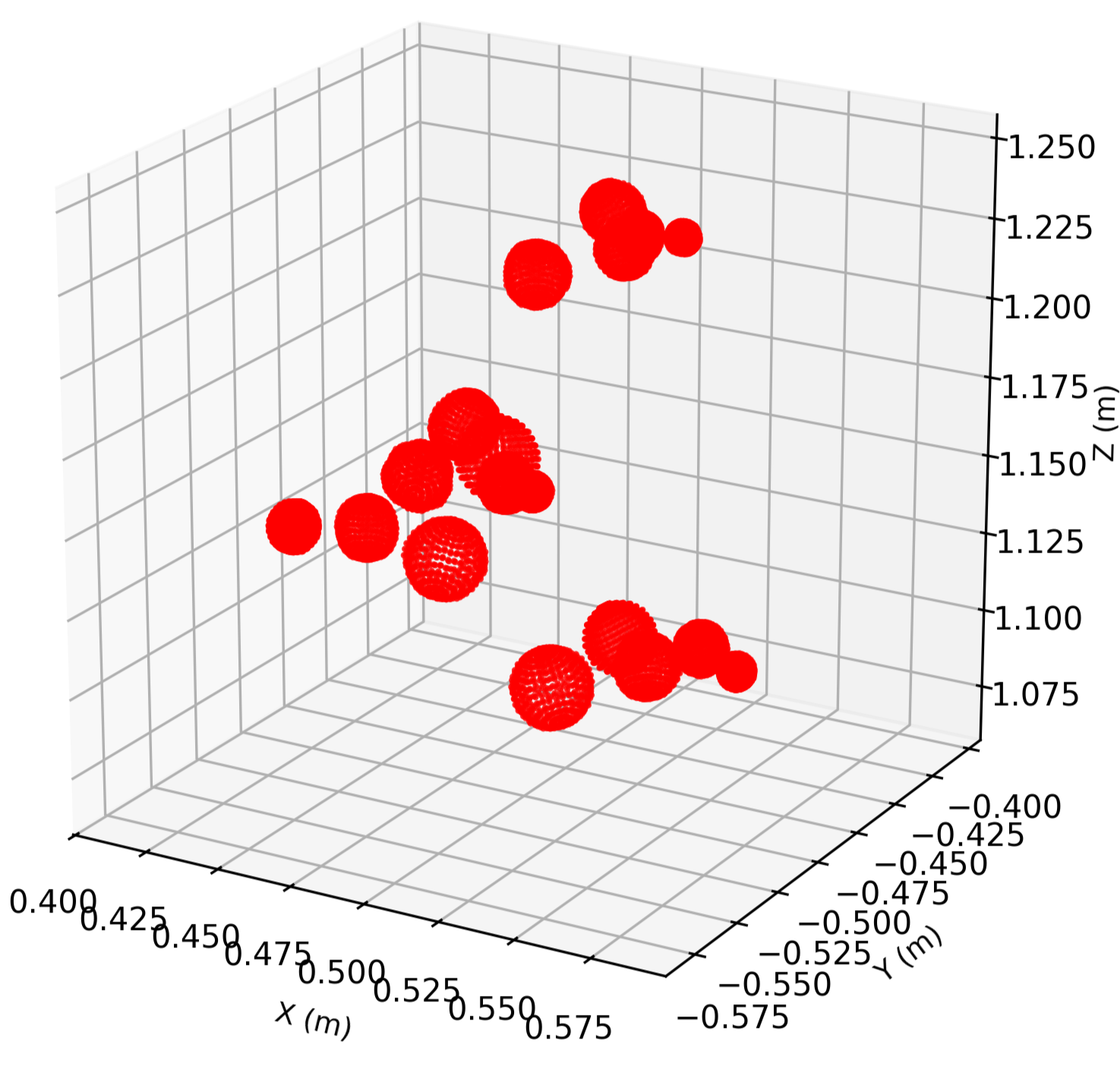}\\[-1mm]
        {\scriptsize }
\includegraphics[width=\linewidth,height=\linewidth]{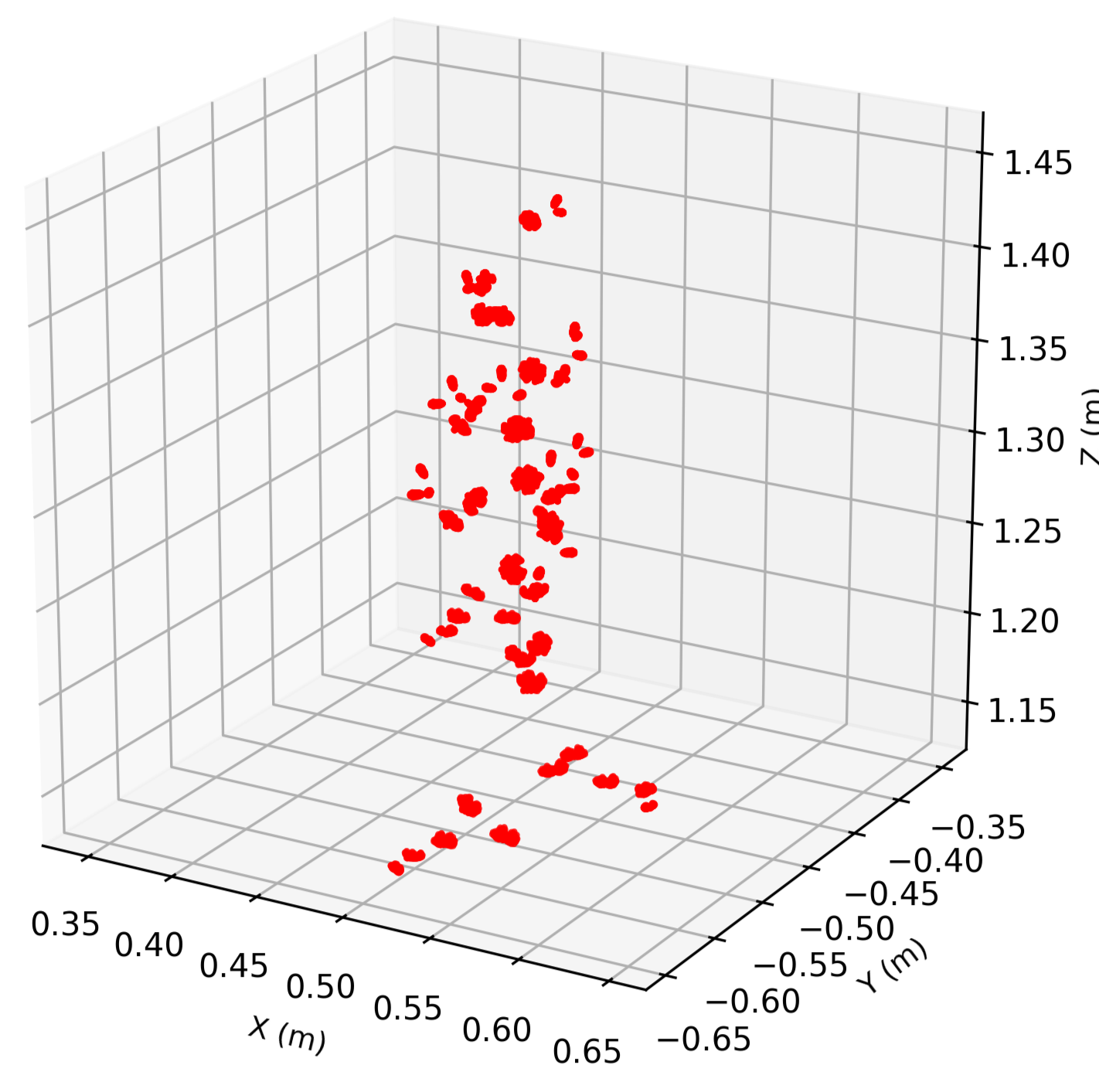}\\[-1mm]
        {\scriptsize  }        \includegraphics[width=\linewidth,height=\linewidth]{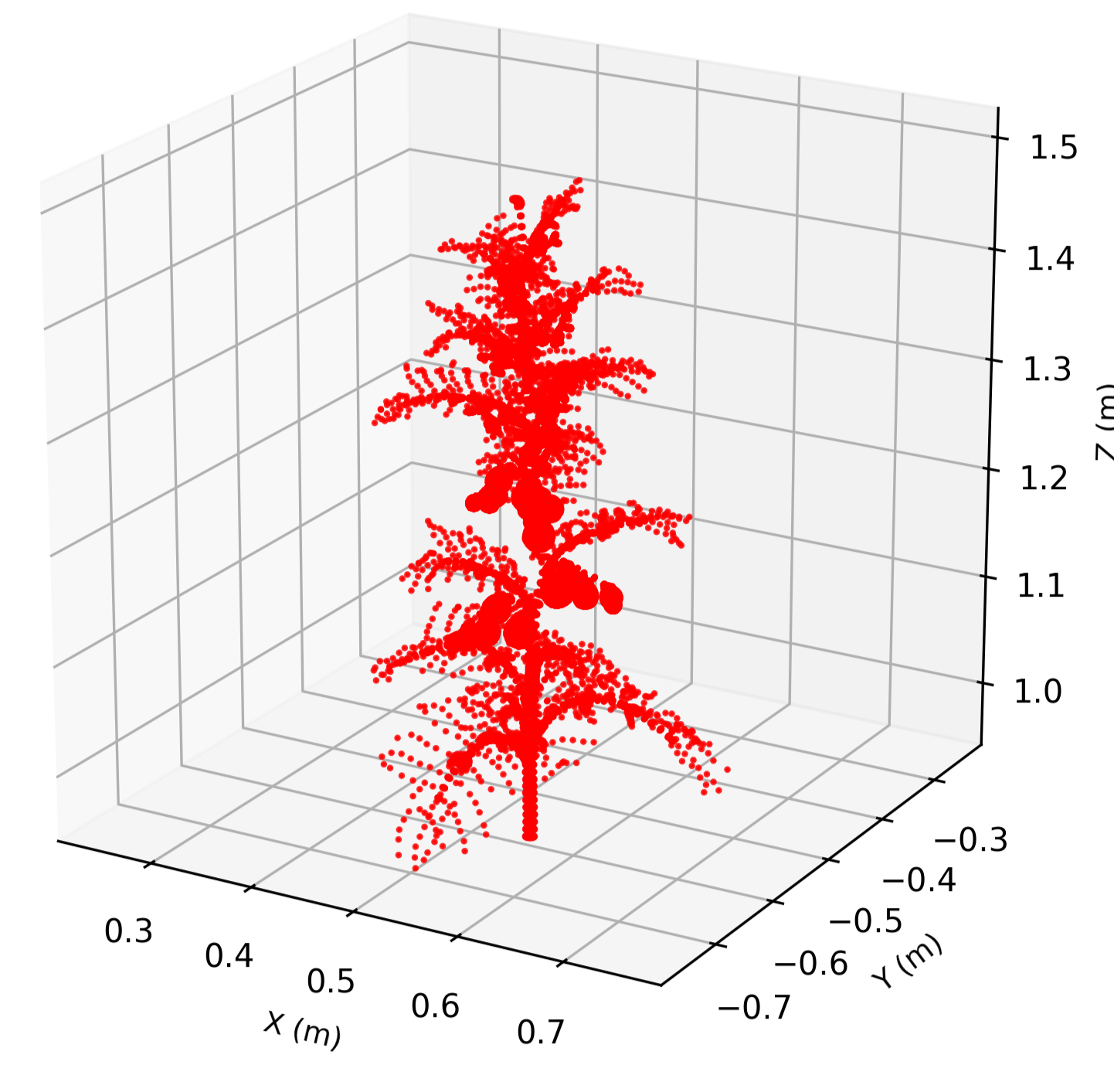}\\[-1mm]
        {\scriptsize Ground Truth.}
    \end{minipage}\hfill
     \begin{minipage}[h]{0.25\columnwidth}
        \centering
\includegraphics[width=\linewidth,height=\linewidth]{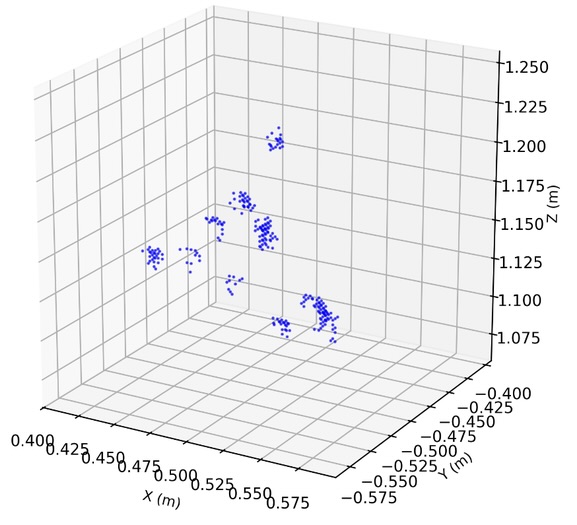}\\[-1mm]
        {\scriptsize }
\includegraphics[width=\linewidth,height=\linewidth]
        {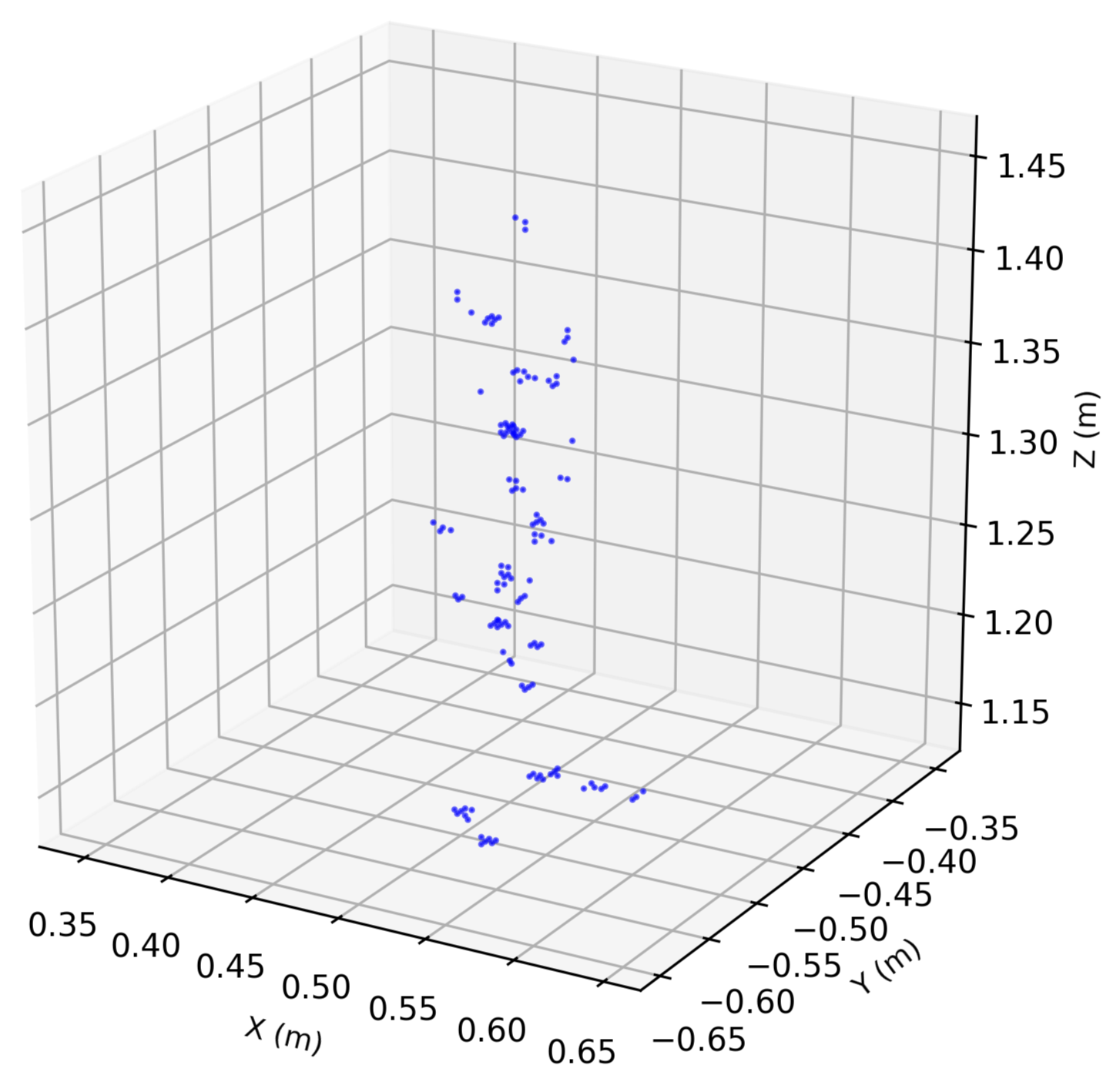}\\[-1mm]
        {\scriptsize }        \includegraphics[width=\linewidth,height=\linewidth]
        {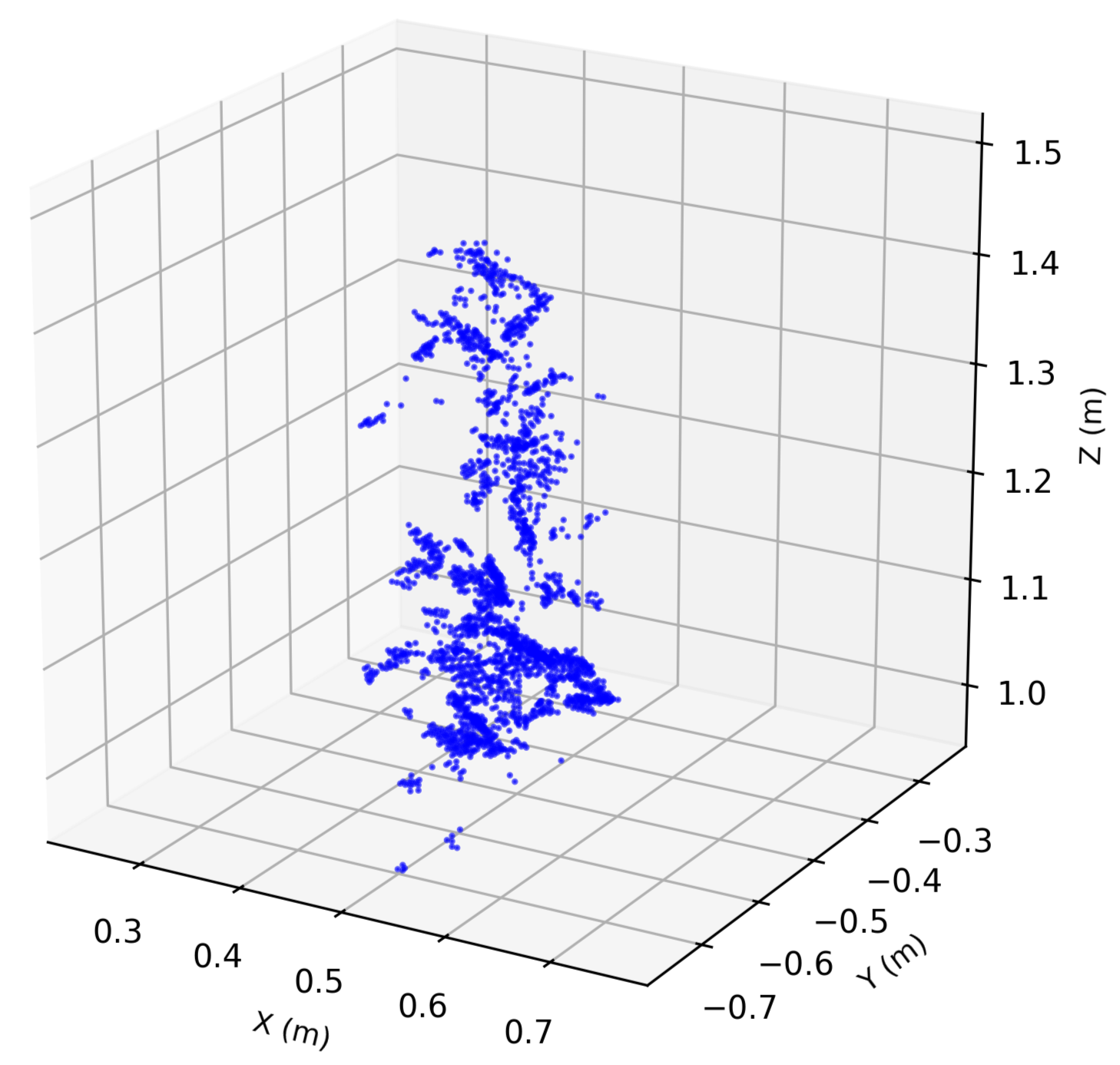}\\[-1mm]
        {\scriptsize View 1}
    \end{minipage}\hfill
     \begin{minipage}[h]{0.25\columnwidth}
        \centering        \includegraphics[width=\linewidth,height=\linewidth]{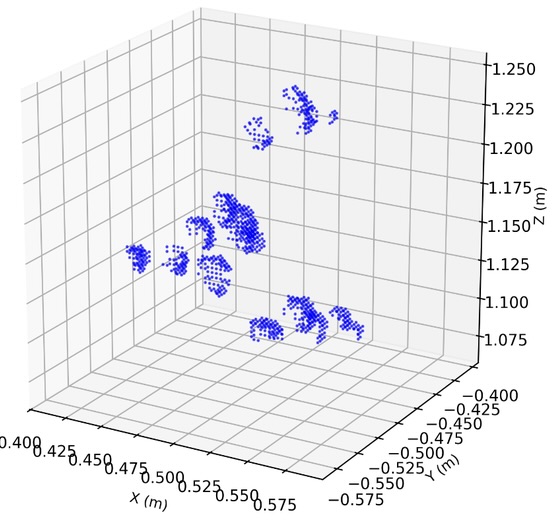}\\[-1mm]
        {\scriptsize }
\includegraphics[width=\linewidth,height=\linewidth]
        {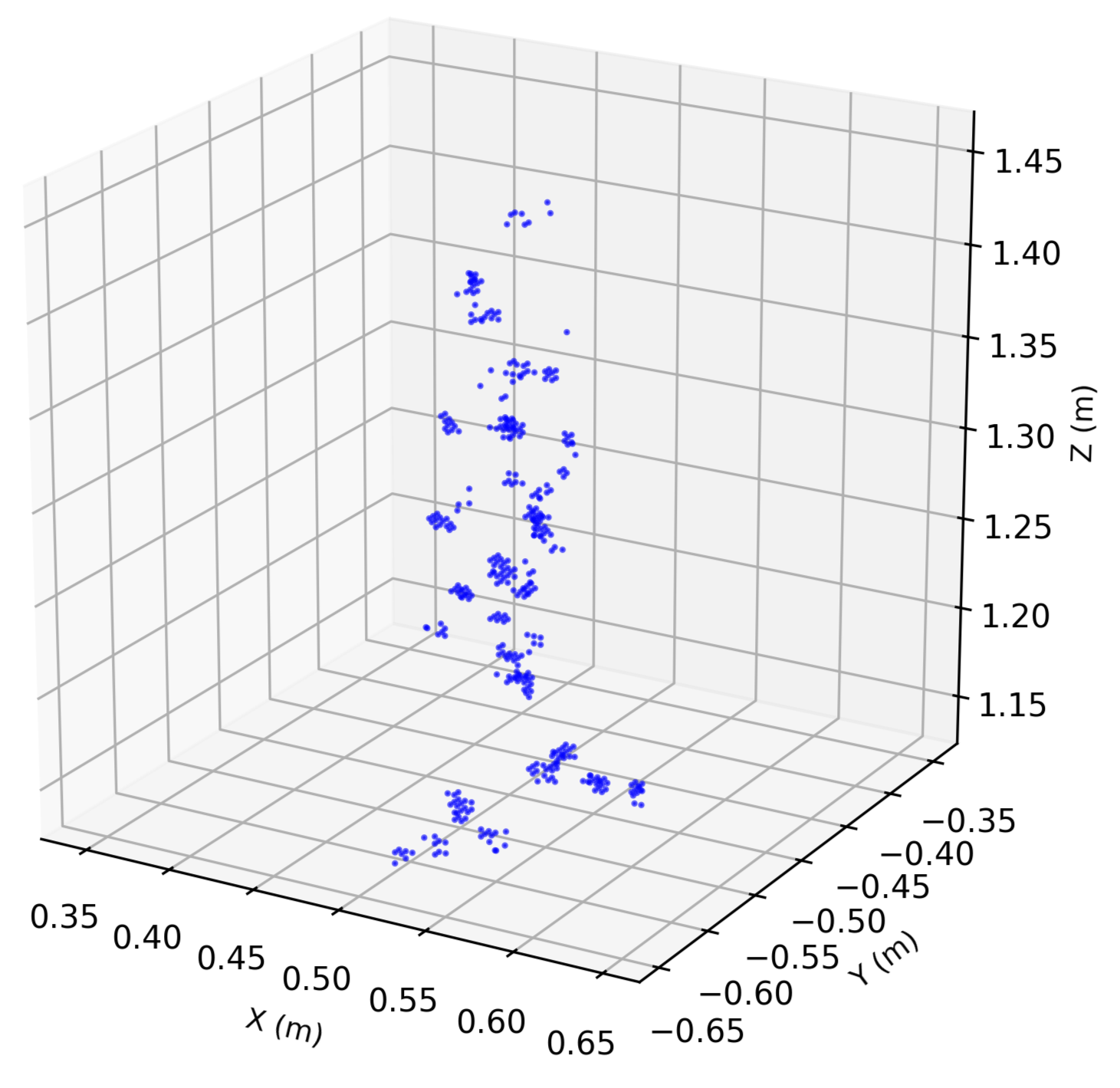}\\[-1mm]
        {\scriptsize }      \includegraphics[width=\linewidth,height=\linewidth]
        {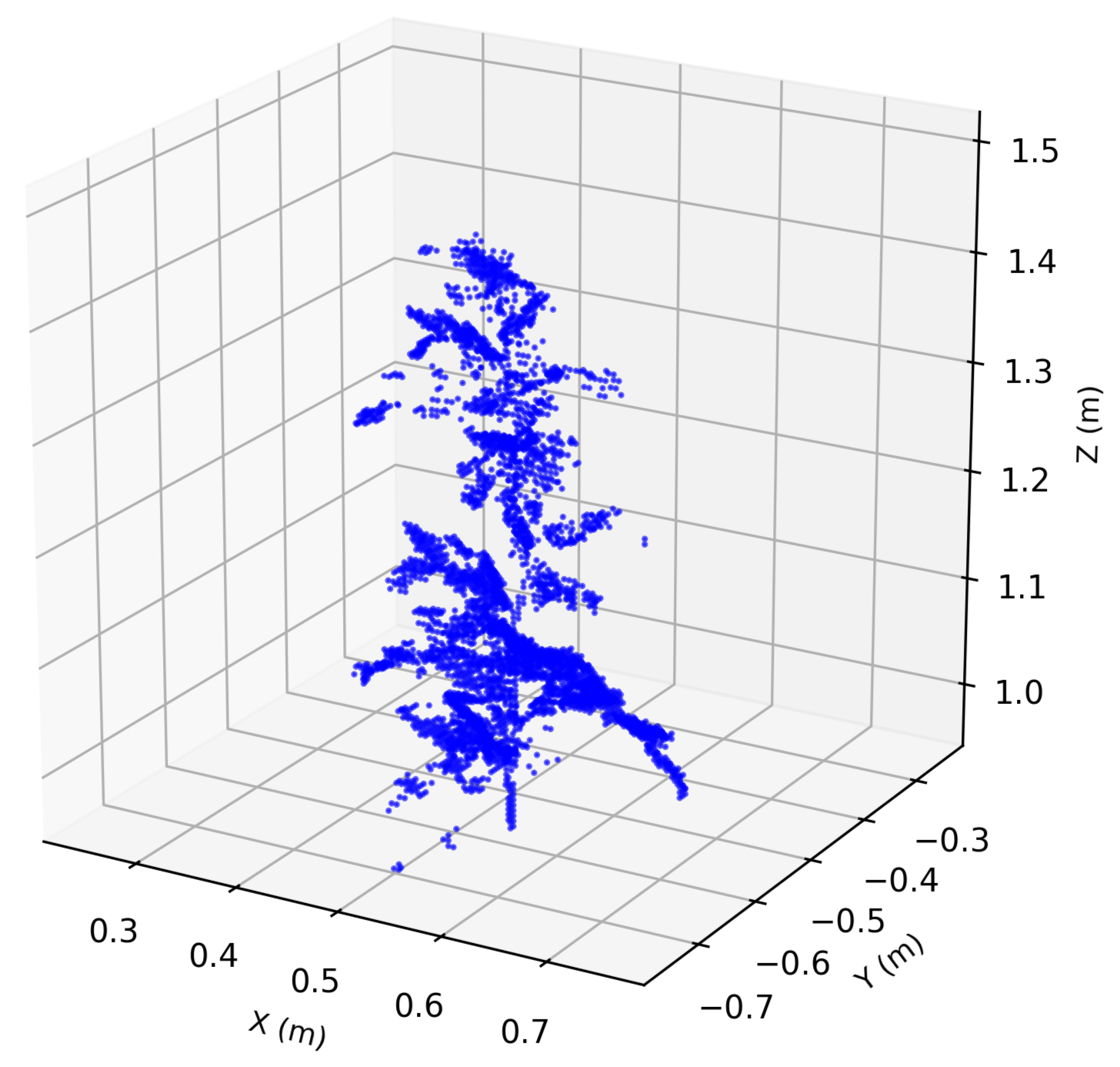}\\[-1mm]
        {\scriptsize View 10}
    \end{minipage}\hfill
     \begin{minipage}[h]{0.25\columnwidth}
        \centering       \includegraphics[width=\linewidth,height=\linewidth]{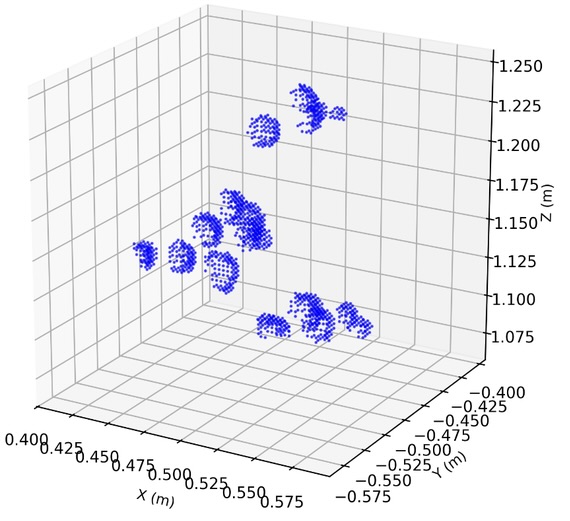}\\[-1mm]
        {\scriptsize}
\includegraphics[width=\linewidth,height=\linewidth]{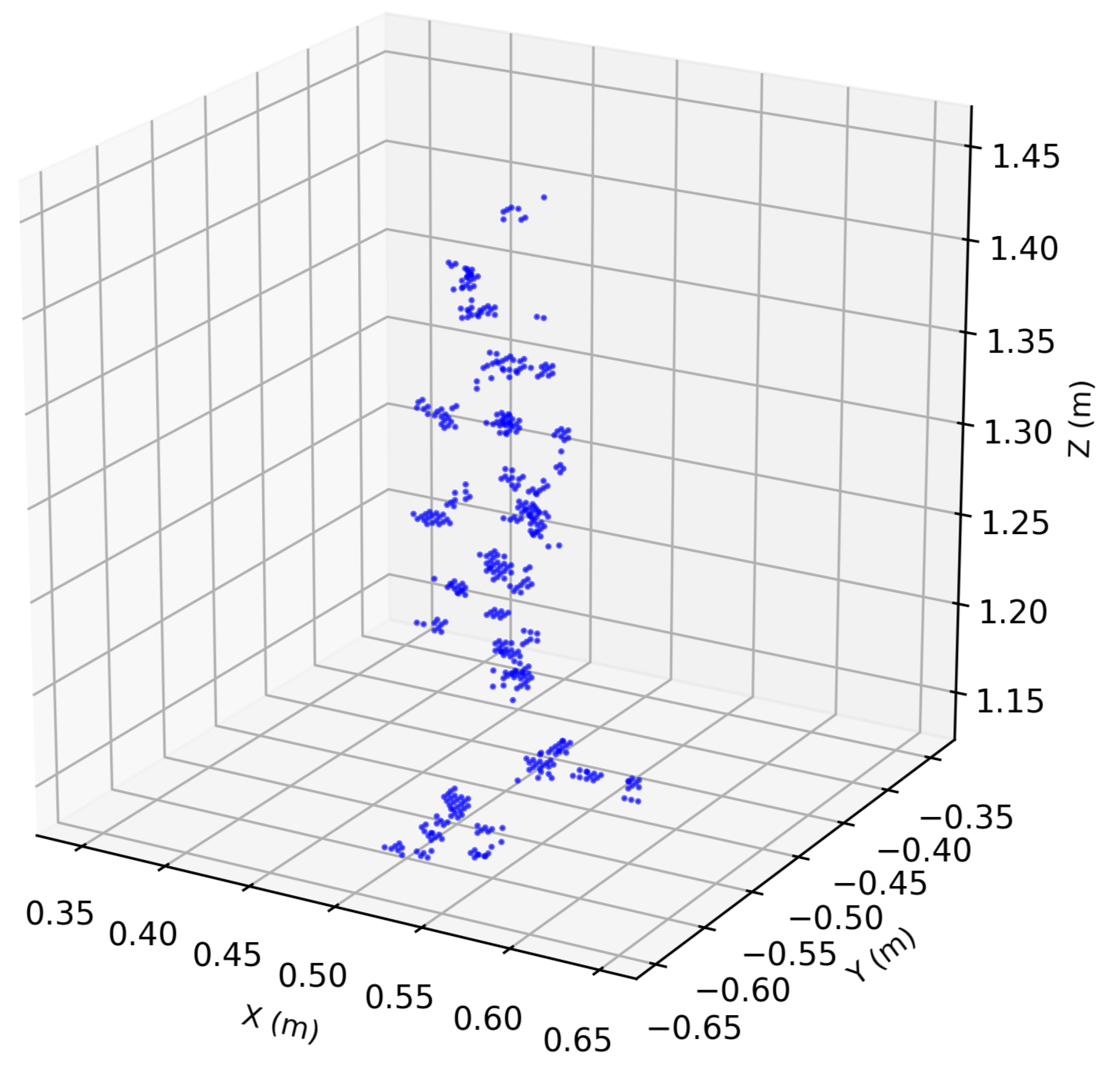}\\[-1mm]
        {\scriptsize }       \includegraphics[width=\linewidth,height=\linewidth]{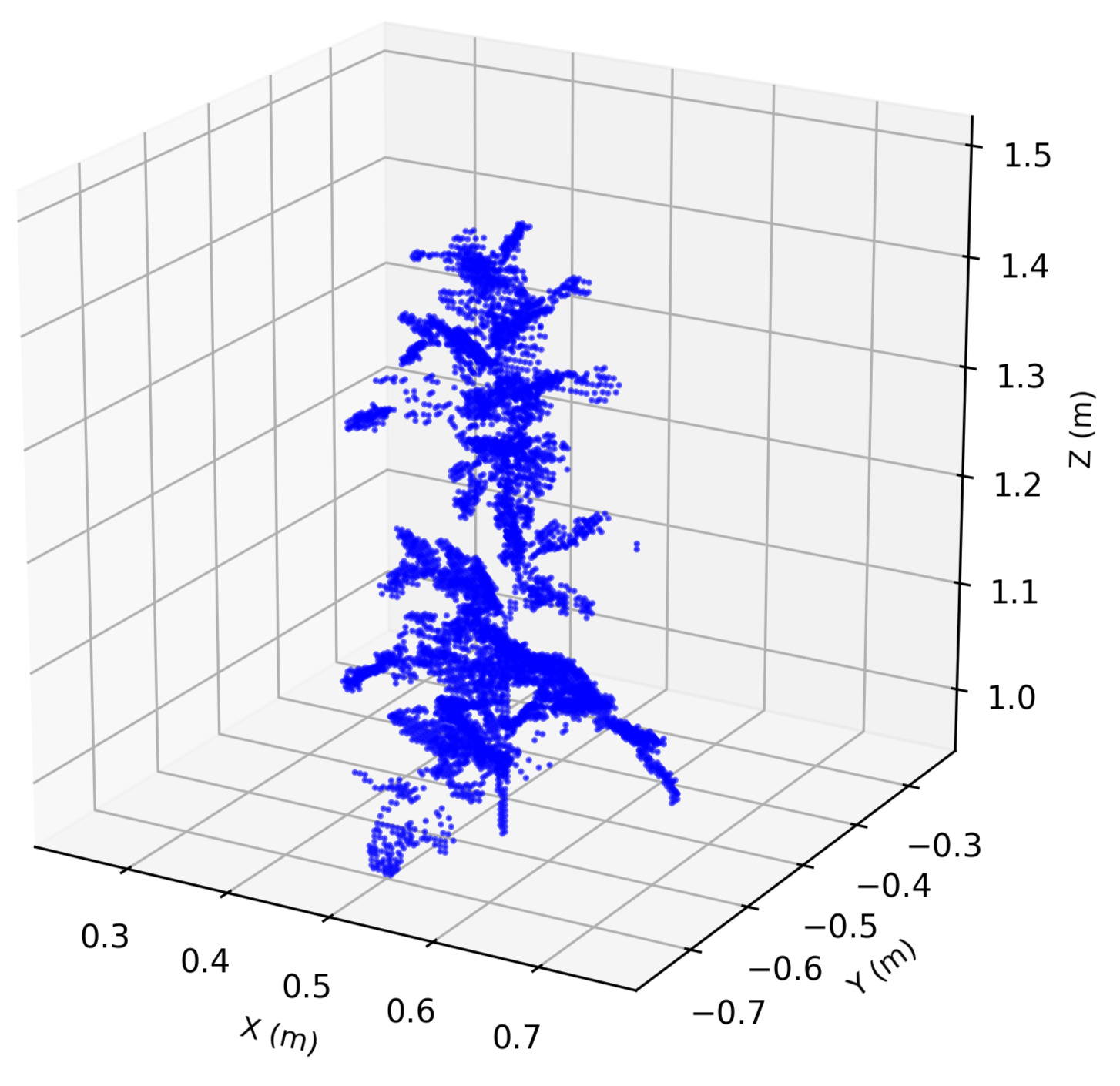}\\[-1mm]
        {\scriptsize View 20}
    \end{minipage}\hfill
    \begin{minipage}[h]{0.25\columnwidth}
        \centering       \includegraphics[width=\linewidth,height=\linewidth]{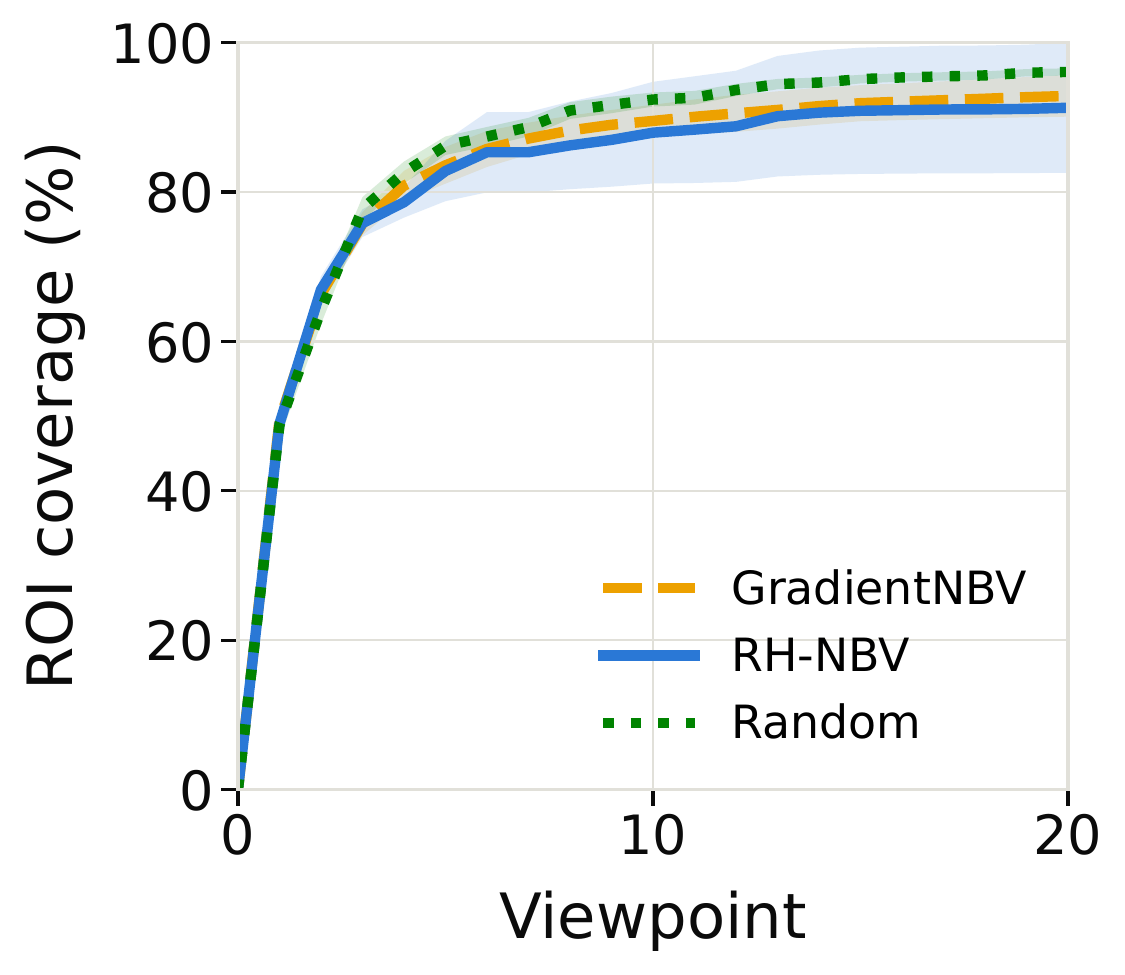}\\[-1mm]
        {\scriptsize}
\includegraphics[width=\linewidth,height=\linewidth]
        {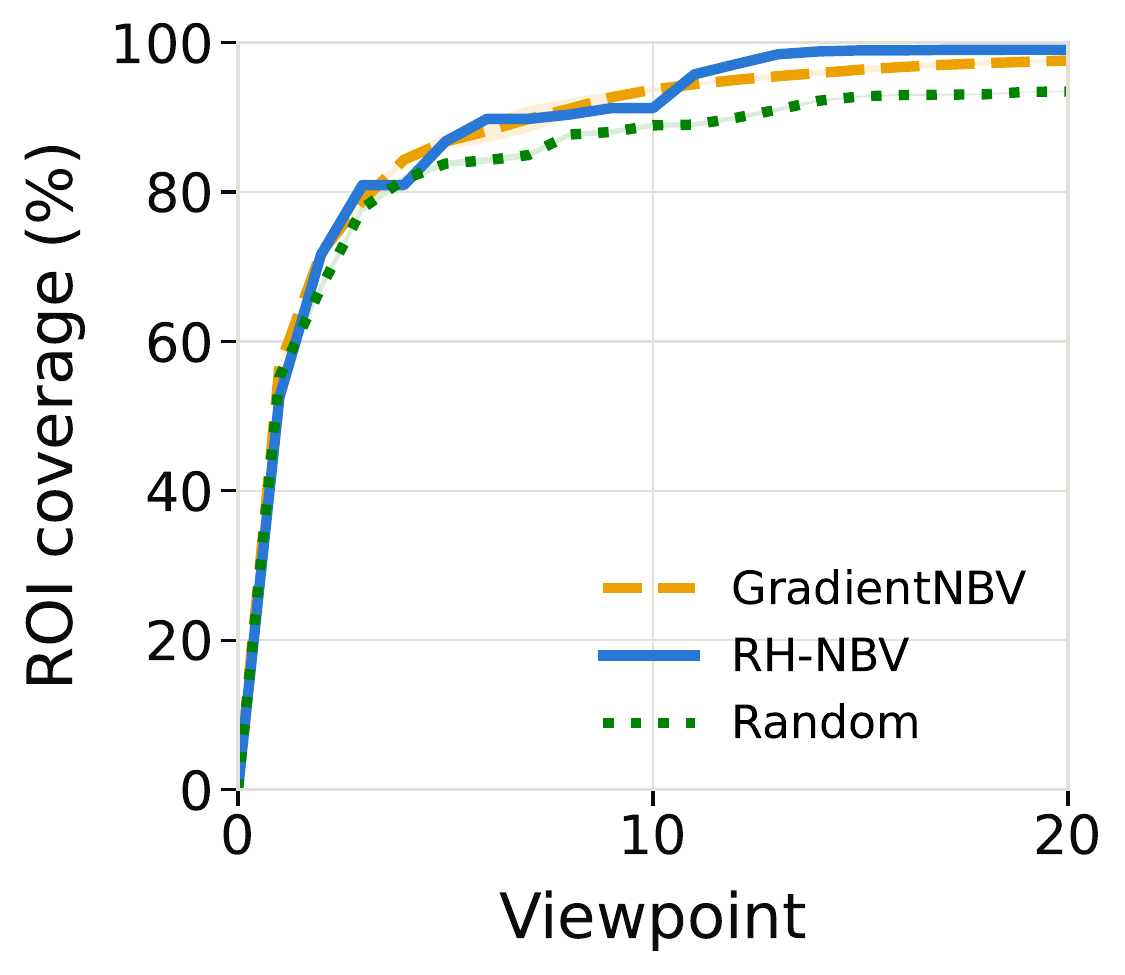}\\[-1mm]
        {\scriptsize }       \includegraphics[width=\linewidth,height=\linewidth]
        {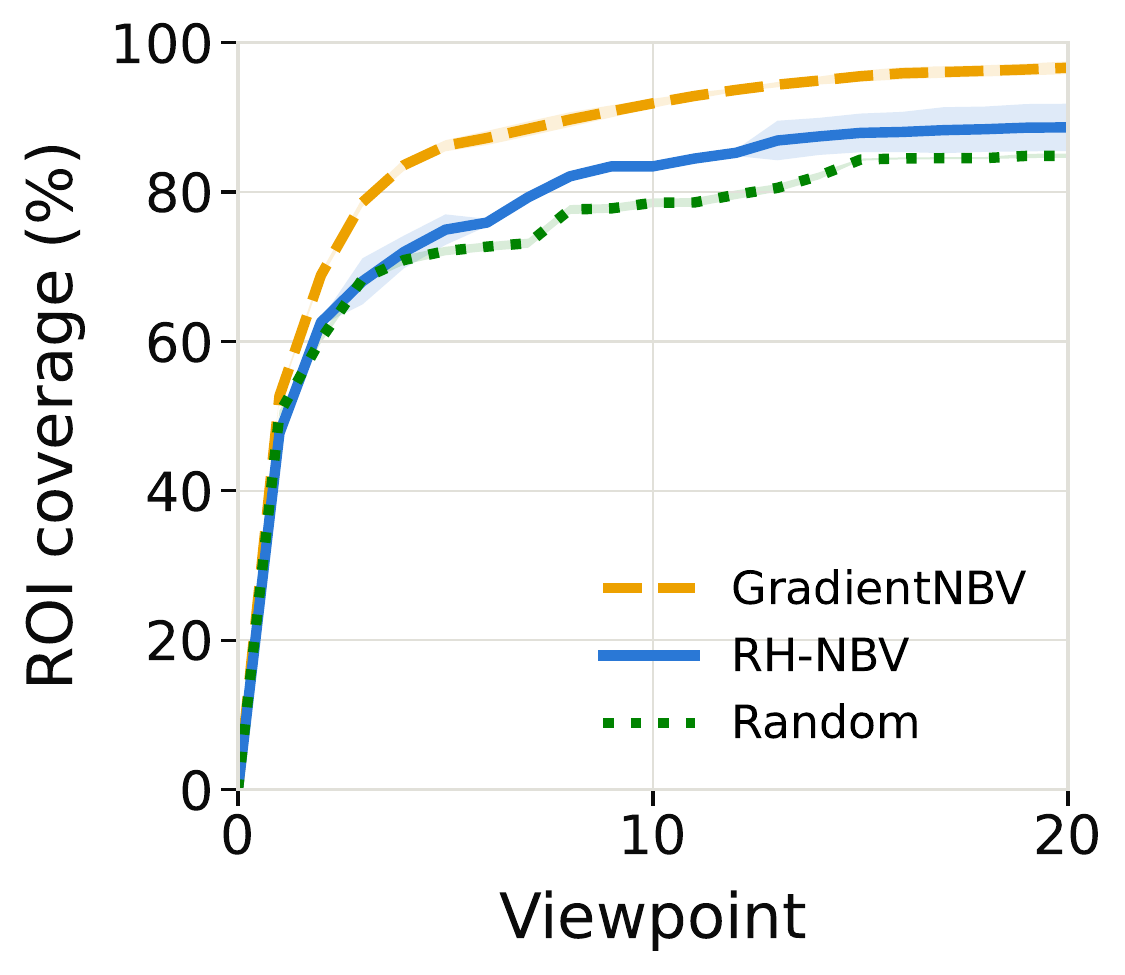}\\[-1mm]
        {\scriptsize ROI Coverage}
    \end{minipage}\hfill
    \begin{minipage}[h]{0.25\columnwidth}
        \centering       \includegraphics[width=\linewidth,height=\linewidth]{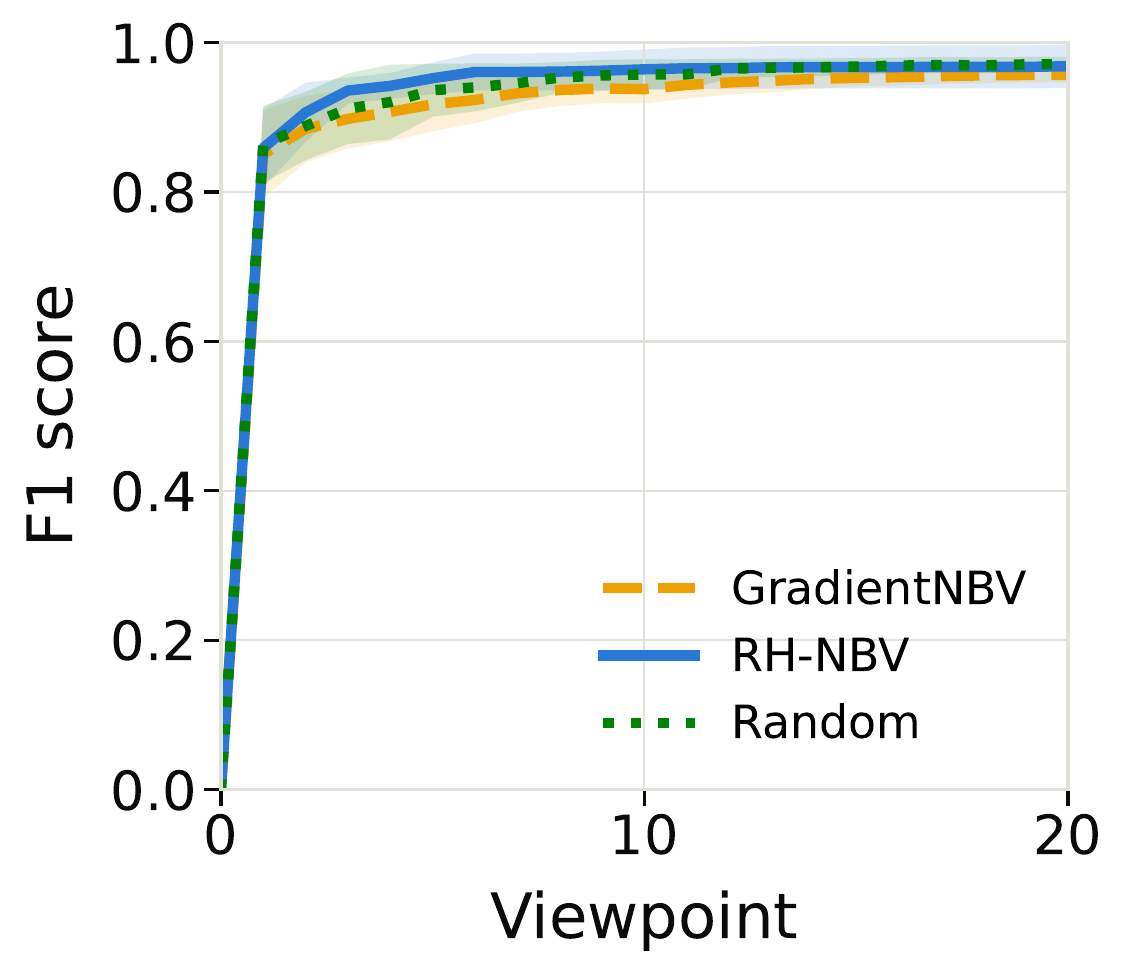}\\[-1mm]
        {\scriptsize}
\includegraphics[width=\linewidth,height=\linewidth]
        {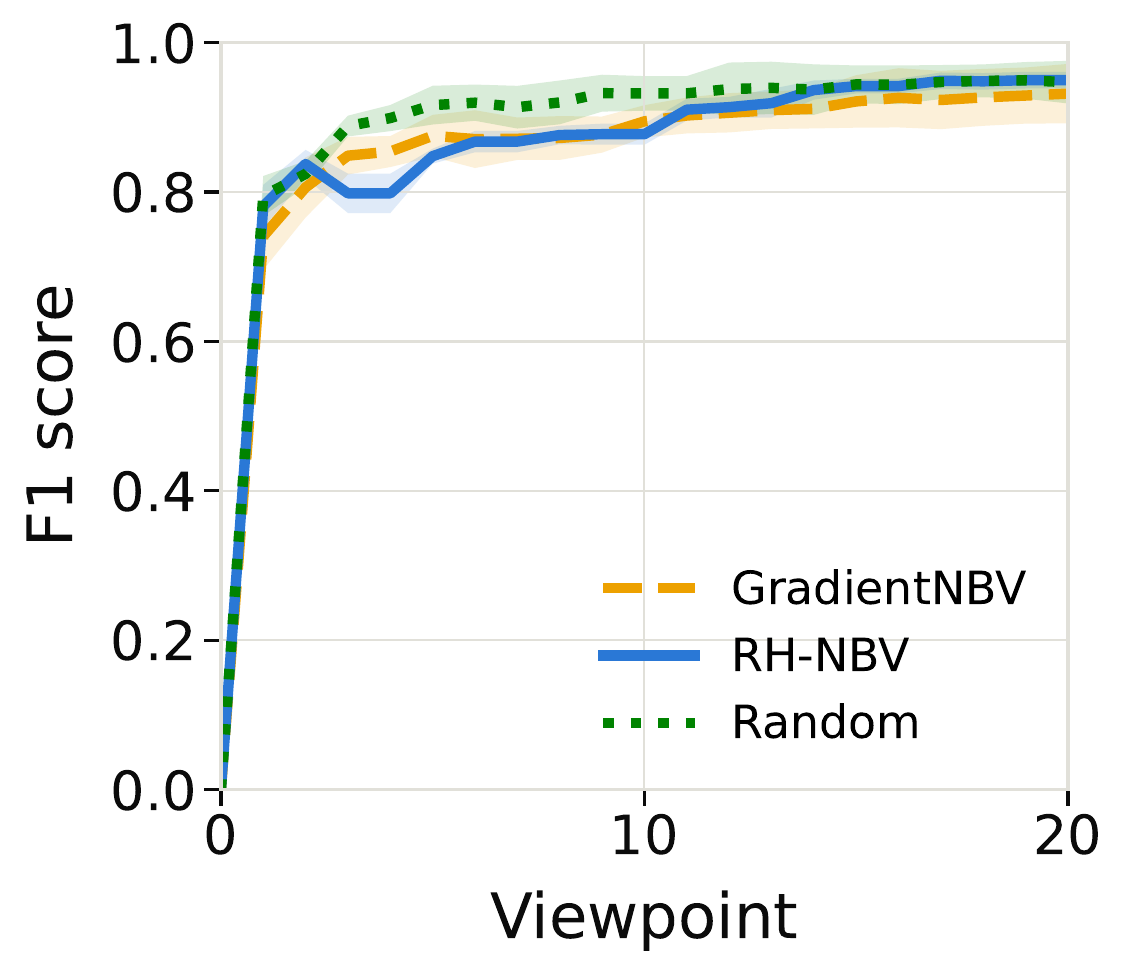}\\[-1mm]
        {\scriptsize }       \includegraphics[width=\linewidth,height=\linewidth]
        {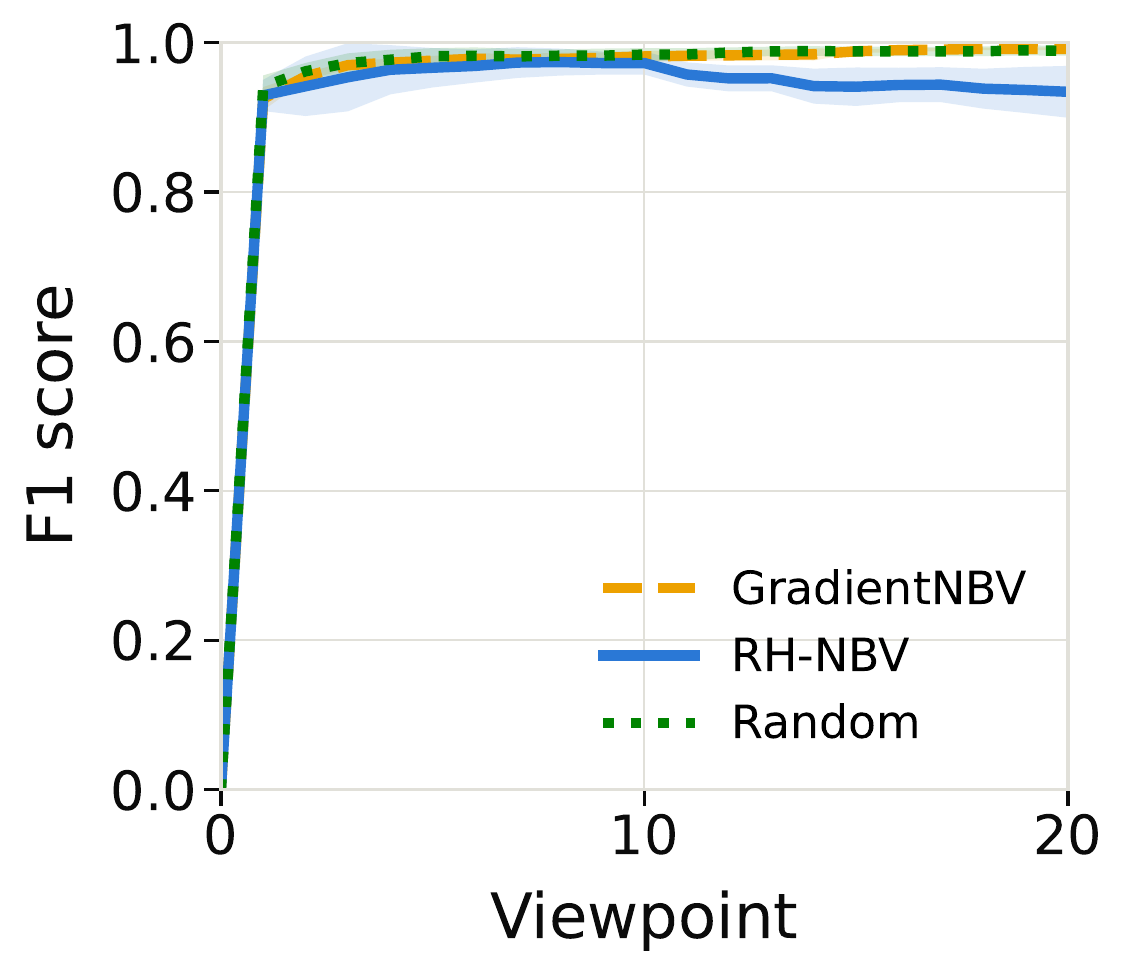}\\[-1mm]
        {\scriptsize F1-score}
    \end{minipage}\hfill
    \caption{Reconstruction performance for the tomato plant at three target attention levels: fruit, blossom, and whole plant. Ground-truth and reconstructed scenes at viewpoints 1, 10, and 20 illustrate progressive target recovery, while the right panels show per-viewpoint ROI coverage and F1 scores for the three planners. RH-NBV is particularly effective for the occlusion-limited fruit and blossom targets, whereas GradientNBV is more competitive for the largely unoccluded whole-plant target.}
\label{fig:tomato-recon}
\end{figure*}

Table~\ref{tab:tomato} reports the results for the tomato plant across three semantic target levels and four plant orientations. Unlike the controlled experiments, no artificial occluders are introduced; visibility is determined solely by the plant geometry. In this setting, GradientNBV achieves the highest final ROI coverage ($95.7\%$), while RH-NBV remains competitive in reconstruction quality, with final $F_1$ scores of $0.970$ and $0.960$, respectively. The myopic planner also requires substantially less computation and motion, using 20 ray-tracing calls and $0.49$~m of trajectory distance compared with 600 calls and $1.77$~m for RH-NBV. Fig.~\ref{fig:tomato-recon} shows the corresponding reconstruction progression and per-viewpoint metrics. 
\begin{table}
\centering
\caption{Comparison of RH-NBV and GradientNBV after 20 executed viewpoints under varying occlusion configurations. Final ROI coverage and occluded-surface recall are reported for the coffee mug and Stanford bunny. $\Delta$ denotes the performance difference of RH-NBV relative to GradientNBV (RH-NBV $-$ GradientNBV).}
\label{tab:ladder}
\resizebox{\linewidth}{!}{%
\begin{tabular}{@{}lccccccc@{}}
\toprule
& \multicolumn{3}{c}{Coverage (\%)} & \multicolumn{3}{c}{Occluded recall} \\
\cmidrule(lr){2-4}\cmidrule(lr){5-7}
Stage & RH-NBV & GradientNBV & $\Delta$ & RH-NBV & GradientNBV & $\Delta$ \\
\midrule
\multicolumn{7}{@{}l}{\emph{Coffee mug}} \\
\texttt{panels1} & 91.59 & 88.84 & $+2.75$ & 0.903 & 0.812 & $+0.091$ \\
\texttt{panels2} & 89.95 & 87.61 & $+2.34$ & 0.868 & 0.799 & $+0.069$ \\
\texttt{panels3} & \textbf{63.82} & 34.23 & $+29.59$ & \textbf{0.855} & 0.074 & $+0.781$ \\
\texttt{panels5} & \textbf{56.93} & 33.96 & $+22.97$ & \textbf{0.712} & 0.070 & $+0.642$ \\
L-shape & \textbf{64.04} & 32.03 & $+32.01$ & \textbf{0.430} & 0.128 & $+0.302$ \\
\midrule
\multicolumn{7}{@{}l}{\emph{Stanford bunny}} \\
\texttt{panels1} & 71.42 & 63.30 & $+8.12$ & 0.839 & 0.733 & $+0.106$ \\
\texttt{panels2} & 53.96 & \textbf{60.63} & $-6.67$ & \textbf{0.719} & 0.684 & $+0.035$ \\
\texttt{panels3} & \textbf{27.32} & 5.66  & $+21.66$ & \textbf{0.309} & 0.021 & $+0.289$ \\
\texttt{{panels5}} & \textbf{31.38} & 4.45  & $+26.93$ & \textbf{0.247} & 0.020 & $+0.227$ \\
L-shape & \textbf{29.67} & 10.46 & $+19.21$ & \textbf{0.235} & 0.136 & $+0.099$ \\
\bottomrule
\end{tabular}%
}
\end{table}
\begin{table*}[t]
\centering
\caption{Results for the occlusion-handling behaviour on the tomato plant under varying level of occlusion. The average performance across 12 trials (3 target attention levels $\times$ 4 plant orientations); all planners share the same predefined initial view (view 0). Lower ray-tracing calls and trajectory distance indicate lower computational and motion cost.}
\label{tab:tomato}
\resizebox{\textwidth}{!}{%
\begin{tabular}{l|ccccc|ccccc|ccccc|ccccc}
\toprule
& \multicolumn{5}{c|}{ROI coverage (\%) $\uparrow$}
& \multicolumn{5}{c|}{$F_1$ score $\uparrow$}
& \multicolumn{5}{c|}{Ray-tracing calls (\#) $\downarrow$}
& \multicolumn{5}{c}{Trajectory distance (m) $\downarrow$} \\
\textbf{\# Viewpoints}
& $0$ & $5$ & $10$ & $15$ & $20$
& $0$ & $5$ & $10$ & $15$ & $20$
& $0$ & $5$ & $10$ & $15$ & $20$
& $0$ & $5$ & $10$ & $15$ & $20$ \\
\midrule
RH-NBV
& 0.0 & 77.3 & 85.6 & 90.7 & 92.8
& 0.000 & 0.927 & \textbf{0.961} & \textbf{0.967} & \textbf{0.970}
& 0 & 150 & 300 & 450 & 600
& 0.00 & 0.36 & 0.64 & 1.32 & 1.77 \\
GradientNBV
& 0.0 & \textbf{85.5} & \textbf{91.7} & \textbf{94.6} & \textbf{95.7}
& 0.000 & 0.923 & 0.938 & 0.954 & 0.960
& 0 & 5 & 10 & 15 & 20
& 0.00 & \textbf{0.18} & \textbf{0.29} & \textbf{0.40} & \textbf{0.49} \\
Random
& 0.0 & 80.7 & 86.6 & 90.8 & 91.5
& 0.000 & \textbf{0.945} & 0.958 & 0.967 & 0.969
& 0 & 0 & 0 & 0 & 0
& 0.00 & 0.49 & 0.95 & 1.45 & 1.75 \\
\bottomrule
\end{tabular}%
}
\end{table*} 
These results complement the controlled-occlusion experiments: when natural self-occlusion leaves the target relatively accessible, immediate viewpoint selection can be sufficient, reducing the benefit of multi-step planning. Thus, RH-NBV is most advantageous when occlusion creates strong dependencies between successive viewpoints, whereas myopic NBV remains preferable when the target can be efficiently exposed through local viewpoint changes.
\subsection{Effect of Planning Horizon}
\label{sec:horizon}
To isolate the effect of non-myopic planning from search effort, we ablate the planning horizon on the coffee mug under \texttt{panels3}, while keeping sampling, utility, and stagnation escape fixed. Table~\ref{tab:ablation} compares different $(H,K)$ configurations.
\begin{table}[]
\centering
\caption{Horizon ablation on the coffee-mug \texttt{panels3} scenario after 20 executed views. Sampling, utility, and stagnation-escape settings are fixed. The results show the effect of planning horizon $H$ and candidate-sequence count $K$ on reconstruction performance and motion cost.}
\label{tab:ablation}
\resizebox{\linewidth}{!}{%
\begin{tabular}{@{}lcccc@{}}
\toprule
Configuration & Cov.\ (\%) & $F_1$ & OccRec & Trajectory distance (m) \\
\midrule
$H{=}3,K{=}10$ (default) & \textbf{63.82} & \textbf{0.892} & \textbf{0.855} & 3.46 \\
$H{=}2,K{=}10$ & 36.97 & 0.500 & 0.354 & 2.33 \\
$H{=}1,K{=}10$ & 36.25 & 0.573 & 0.443 & 2.75 \\
$H{=}1,K{=}30$ (equal compute) & 34.51 & 0.198 & 0.113 & 0.64 \\
\bottomrule
\end{tabular}%
}
\end{table}
\begin{figure}[]
\centering
\includegraphics[width=\linewidth]{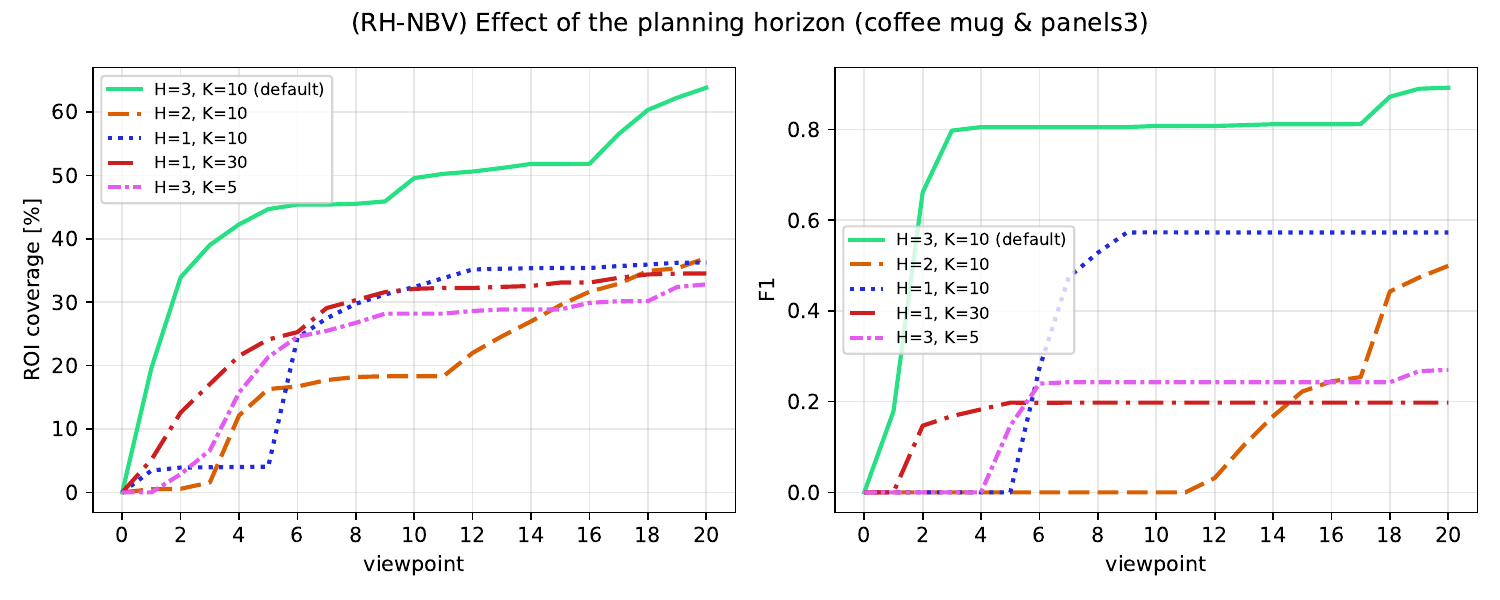}
\caption{ROI coverage (left) and $F_1$ (right) vs.\ viewpoint index for the configurations of Table~\ref{tab:ablation}. The default ($H{=}3,K{=}10$) separates from every other setting within the first few viewpoints and keeps gaining, while the reduced-horizon settings plateau between 33--37\% coverage.}
\label{fig:horizon}
\end{figure}
As Table~\ref{tab:ablation} shows, the default $H=3,K=10$ substantially outperforms reduced horizons ($H=2$ or $H=1$), and increasing the number of single-step candidates does not recover this performance: the equal-compute configuration $H=1,K=30$ performs even worse than $H=1,K=10$. Thus, evaluating more viewpoints alone is insufficient; the key difference is that RH-NBV evaluates viewpoints as coupled sequences and propagates the predicted state between successive observations. Fig.~\ref{fig:horizon} shows the resulting separation in coverage and $F_1$ throughout the 20-viewpoint sequence, supporting that the performance gain arises from the planning horizon itself rather than from increased candidate evaluation.
 \subsection{Performance--Cost Trade-off}
\label{sec:cost}

The performance gains of RH-NBV come with additional computational and motion cost. With $K=10$ and $H=3$, RH-NBV evaluates 30 hypothetical viewpoints per iteration, resulting in 600 ray-tracing calls over 20 executed viewpoints, compared with 20 for GradientNBV and none for Random. It also travels farther, reaching $2.15$~m on the mug and $1.50$~m on the bunny, compared with $0.50$~m and $0.42$~m for GradientNBV, respectively. Thus, RH-NBV trades additional computation and motion for improved recovery of difficult-to-observe target regions. The progressive gains reported in Table~\ref{tab:mug-bunny-burusa} further indicate that this benefit accumulates throughout the reconstruction sequence rather than appearing only in the final view.

\subsection{Real-World Experiment}
\label{sec:real-robot}

RH-NBV ($K{=}10$, $H{=}3$) was evaluated on a physical UR5e equipped with an eye-in-hand Intel RealSense D455 and compared with GradientNBV and Random on a coffee mug under two occlusion configurations. The easy configuration places panels behind and to the right of the mug, whereas the hard configuration encloses the target with four panels (Fig.~\ref{fig:physical_setup}). Each planner was evaluated after 10 executed viewpoints; results are reported in Table~\ref{tab:real}. 

Under easy occlusion, GradientNBV achieves the highest ROI coverage ($99.24\%$), whereas RH-NBV obtains the highest reconstruction $F_1$ score ($0.367$ vs.~$0.180$ for GradientNBV and $0.318$ for Random). Under hard occlusion, RH-NBV achieves both the highest ROI coverage ($90.24\%$) and $F_1$ score ($0.222$), outperforming GradientNBV ($78.64\%$, $0.040$) and Random ($87.96\%$, $0.018$). The same overall trend was observed in the simulation experiments, where the advantage of RH-NBV became more pronounced as occlusion increased.  

\begin{table}[H]
\centering
\caption{Performance on Real-robot experiments.}
\label{tab:real}
\resizebox{\linewidth}{!}{%
\begin{tabular}{@{}llccc@{}}
\toprule
Scenario & Method & Cov.\ (\%) & $F_1$ & Trajectory distance (m)  \\
\midrule
\multirow{3}{*}{Easy }
 & RH-NBV      & 83.84 & 0.367 & 1.17  \\
 & GradientNBV & 99.24 & 0.180 & 0.35  \\
 & Random      & 90.13 & 0.318 & 0.92  \\
\midrule
\multirow{3}{*}{Hard }
 & RH-NBV      & 90.24 & 0.222 & 1.15  \\
 & GradientNBV & 78.64 & 0.040 & 0.26  \\
 & Random      & 87.96 & 0.018 & 0.93  \\
\bottomrule   
\end{tabular}%
}
\end{table}

\section{Discussion}
\label{sec:discussion}
Overall, the experiments show that the advantage of RH-NBV emerges primarily under strong target occlusion. In the controlled occlusion experiments, RH-NBV achieves substantially higher occluded-target recall in the more challenging configurations, while its advantage is smaller when the target is more visible. The tomato-plant experiment shows the same trend within a single scene: RH-NBV is more effective gaining information in occluded fruits, whereas GradientNBV becomes preferable for the largely unoccluded whole-plant target. The horizon ablation further indicates that this improvement comes from multi-step planning rather than simply evaluating more viewpoints, as $H=3,K=10$ outperforms the matched-evaluation-budget $H=1,K=30$ case. These results suggest that RH-NBV is most useful when occlusion creates sequential dependencies between informative viewpoints; when such dependencies are weak, myopic planning remains a simpler and more computationally efficient choice. 

\section{Conclusion}
\label{sec:conclusion}

This paper presented RH-NBV, a receding-horizon next-best-view planner for acquiring information about occluded targets. The key idea is to move beyond myopic viewpoint selection by evaluating short sequences of future views while propagating a predicted belief state, thereby capturing the sequential dependence between observations. Across controlled external and natural self-occlusion experiments, RH-NBV provided its greatest benefit when target visibility was strongly constrained, improving recovery of otherwise difficult-to-observe target regions, while myopic NBV remained competitive when the target was more readily visible. Horizon ablations further showed that this improvement resulted from multi-step reasoning rather than simply evaluating a larger number of single-step candidates. Real-robot experiments on a UR5e with an eye-in-hand RGB-D camera demonstrated that the approach transfers to physical hardware, with the improved perception performance accompanied by higher computational and motion cost. Future work will investigate adaptive planning horizons that adjust the extent of look-ahead according to the estimated visibility and uncertainty of the target, with the goal of retaining the benefits of non-myopic planning while reducing its computational overhead.

\typeout{get arXiv to do 4 passes: Label(s) may have changed. Rerun}


\begin{thebibliography}{00}


\bibitem{kootstra_wang_blok_hemming_henten_2021}
G. Kootstra, X. Wang, P. M. Blok, J. Hemming, and E. van Henten, ``Selective harvesting robotics: Current research, trends, and future directions,'' \textit{Current Robotics Reports}, vol.~2, pp.~95--104, Jan. 2021.

\bibitem{bajcsy_1988}
R. Bajcsy, ``Active perception,'' \textit{Proceedings of the IEEE}, vol.~76, no.~8, pp.~966--1005, 1988.

\bibitem{aloimonos_weiss_bandyopadhyay_1988}
J. Aloimonos, I. Weiss, and A. Bandyopadhyay, ``Active vision,'' \textit{International Journal of Computer Vision}, vol.~1, no.~4, pp.~333--356, Jan. 1988.

\bibitem{connolly_1985}
C. Connolly, ``The determination of next best views,'' in \textit{Proc. 1985 IEEE Int. Conf. on Robotics and Automation}, Mar. 1985.

\bibitem{zaenker_claus_mccool_bennewitz_2020}
T. Zaenker, C. Smitt, C. McCool, and M. Bennewitz, ``Viewpoint planning for fruit size and position estimation,'' arXiv:2011.00275, 2020.
 

\bibitem{bircher_kamel_alexis_oleynikova_siegwart_2016}
A. Bircher, M. Kamel, K. Alexis, H. Oleynikova, and R. Siegwart, ``Receding horizon `next-best-view' planner for 3D exploration,'' in \textit{Proc. IEEE Int. Conf. on Robotics and Automation (ICRA)}, May 2016, pp.~1462--1468.

\bibitem{zhang_zhang_2022}
H. Zhang and Z. Zhang, ``Volumetric information gain guided receding horizon planner for active 3D reconstruction,'' in \textit{2022 IEEE Int. Conf. on Robotics and Biomimetics (ROBIO)}, Dec. 2022, pp.~1233--1238.

\bibitem{delmerico2018}
J. Delmerico, S. Isler, R. Sabzevari, and D. Scaramuzza, ``A comparison of volumetric information gain metrics for active 3D object reconstruction,'' \textit{Autonomous Robots}, vol.~42, no.~2, pp.~197--208, 2018.


\bibitem{burusa_henten_gert_kootstra_2024}
A. K. Burusa, E. J. van Henten, and G. Kootstra, ``Gradient-based local next-best-view planning for improved perception of targeted plant nodes,'' in \textit{Proc. 2024 IEEE Int. Conf. on Robotics and Automation (ICRA)}, Yokohama, Japan, May 2024, pp.~15854--15860.

\end{thebibliography}
\end{document}